\documentclass{article}
\usepackage{graphicx} 
\usepackage{geometry}

\usepackage{pdflscape}

\usepackage{booktabs, multirow} 
\usepackage{soul}
\usepackage[table, dvipsnames]{xcolor} 
\usepackage{tabularx}
\usepackage{changepage,threeparttable} 
\usepackage{multicol}
\usepackage{verbatim}
\usepackage{ltablex}
\usepackage{xltabular}
\usepackage{booktabs}
\usepackage{authblk}

\usepackage{enumitem}
\usepackage{wrapfig}
\usepackage[most]{tcolorbox}

\usepackage{tabularx} 
\usepackage{booktabs} 

\newenvironment{WrapText}[1][r]
  {\wrapfigure{#1}{0.53\textwidth}\tcolorbox}
  {\endtcolorbox\endwrapfigure}

\usepackage{pdfpages}

\title{Framework-Based Qualitative Analysis of Free Responses of Large Language Models: Algorithmic Fidelity}

\author[1]{Aliya Amirova}
\author[2]{Theodora Fteropoulli}
\author[3]{Nafiso Ahmed}
\author[4,5]{Martin R. Cowie}
\author[6]{Joel Z. Leibo\small}

\affil[1]{\small Population Health Sciences,
School of Life Course \&  Population Sciences, Faculty of Life Sciences \& Medicine, King’s College London, UK}
\affil[2]{Medical School, University of Cyprus, Nicosia, Cyprus}
\affil[3]{Division of Psychiatry, University College London, UK}

\affil[4]{Royal Brompton Hospital, London, UK}

\affil[5]{School of Cardiovascular Medicine \& Sciences, Faculty of Life Sciences \& Medicine, King’s College London, UK}

\affil[6]{Google DeepMind, London, UK}
\affil[7]{Department of Informatics, Faculty of Natural, Mathematical \& Engineering Sciences, King’s College London, UK}

\begin{document}

\begingroup
\let\center\flushleft
\let\endcenter\endflushleft
\maketitle
\endgroup

  
\begin{abstract}

Today, with the advent of Large-scale generative Language Models (LLMs) it is now possible to simulate free responses to interview questions such as those traditionally analyzed using qualitative research methods. Qualitative methodology encompasses a broad family of techniques involving manual analysis of open-ended interviews or conversations conducted freely in natural language. Here we consider whether artificial ``silicon participants'' generated by LLMs may be productively studied using qualitative analysis methods in such a way as to generate insights that could generalize to real human populations. The key concept in our analysis is \emph{algorithmic fidelity}, a validity concept capturing the degree to which LLM-generated outputs mirror human sub-populations' beliefs and attitudes. By definition, high algorithmic fidelity suggests that latent beliefs elicited from LLMs may generalize to real humans, whereas low algorithmic fidelity renders such research invalid. Here we used an LLM to generate interviews with ``silicon participants'' matching specific demographic characteristics one-for-one with a set of human participants. Using framework-based qualitative analysis, we showed the key themes obtained from both human and silicon participants were strikingly similar. However, when we analyzed the structure and tone of the interviews we found even more striking differences. We also found evidence of a hyper-accuracy distortion. We conclude that the LLM we tested (GPT-3.5) does not have sufficient algorithmic fidelity to expect \textit{in silico} research on it to generalize to real human populations. However, rapid advances in artificial intelligence raise the possibility that algorithmic fidelity may improve in the future. Thus we stress the need to establish epistemic norms now around how to assess the validity of LLM-based qualitative research, especially concerning the need to ensure the representation of heterogeneous lived experiences.

\end{abstract}
\newpage

\section*{Introduction}

\begin{WrapText}
{\large \textbf{Glossary}}
{\footnotesize
\begin{itemize}[leftmargin=*]

\item \textbf{Large language model (LLM):} a conditional distribution over the next word in a document given the sequence of words leading up to it.

\item \textbf{Algorithmic fidelity:} the degree to which beliefs and attitudes sampled from an appropriately conditioned LLM accurately reflect those of real human subpopulations, i.e.~a validity concept used to evaluate when research on human behavior using LLMs to generate data may be expected to generalize.

\item \textbf{Silicon participant:} LLM-generated output conditioned on contextual information such as age, gender, etc, chosen to match a human participant. 

\item \textbf{Silicon sampling} is the process of simulating a population of silicon participants using context to define subpopulations. If there is sufficient algorithmic fidelity then the simulated subpopulations so produced closely track the attitudes, opinions, and experiences of their corresponding human subpopulations. 


\item \textbf{Public-patient engagement:} actively soliciting knowledge, values, and norms from stakeholders to create a shared perspective for conducting public health and social care research, i.e.~research carried out ‘with’ or ‘by’ stakeholders rather than ‘to’, ‘about’ or ‘for’ them, following principles of respect, support, transparency, responsiveness, fairness of opportunity, and accountability. 

\item \textbf{Semi-structured interview:} a versatile way of collecting qualitative data (i.e.~free-form text) using an interview guide tailored to the research aim and based on prior work, theory, or a framework.

\item \textbf{Framework-based qualitative analysis:} an approach to qualitative analysis that finds patterns in free-form text, examines their relationships and derives common themes.

\item \textbf{Theoretical construct:} a TDF term referring to a component part of a theory hypothesised by that theory to predict human behaviour. 

\item \textbf{Theoretical domain:} a TDF term referring to a grouping of theoretical constructs produced by expert consensus.

\item \textbf{Belief statement:} a summary of a set of `quotes' (e.g.~words, sentences, collections of sentences) that share a common meaning. Belief statements are in one-to-one correspondence with unique meanings.
\end{itemize}
}
\end{WrapText}

Large-Scale generative Language Models (LLMs) \cite{brown2020language, openai2023gpt4, anil2023palm} may provide a new opportunity for capturing available knowledge and beliefs at scale and facilitate \emph{in silico} research on human behavior and cognition \cite{argyle2023out, brand2023using, aher2023using, dillion2023can, grossmann2023ai}. By virtue of their training, LLMs may contain substantial latent social information \cite{horton2023large}---enough to consider them plausible computational models of humans. LLMs may capture economic laws, decision-making heuristics, and social preferences \cite{horton2023large}, as well as mirroring human moral judgments \cite{dillion2023can}. In principle, using LLMs could be a cost-effective and efficient way to gain insights and explore how self-reports vary, as well as to pilot experiments \emph{in silico} to test sensitivity of responses to precise wording. Cheap and easy to run \emph{in silico} experiments can guide expensive and slow empirical work with real participants. However, how can we know such results are trustworthy? 

\textit{Algorithmic fidelity} is an appropriate validity concept for research on human behavior using survey data simulated using Large-scale generative Language Models (LLMs) \cite{argyle2023out}. Algorithmic fidelity describes the extent to which the outputs of LLMs conditioned to simulate specific human sub-populations actually reflect the beliefs and attitudes of those subpopulations.

By definition, high algorithmic fidelity suggests that beliefs elicited from the LLM will generalize to real humans, while low fidelity renders such inferences invalid. Argyle et al. \cite{argyle2023out} introduced the idea of generating \emph{``silicon samples''} or \emph{``silicon participants''} by conditioning LLMs using backstories matching sociodemographic data from human survey participants. Their study found that LLM outputs closely mirrored the complex interplay of beliefs, attitudes, and sociocultural contexts that shape human responses to surveys about American politics. In the present study, we aim to extend algorithmic fidelity assessment methodology \cite{argyle2023out} using qualitative research methods so that it can be applicable to studies where the data consists of fully freeform responses in natural language.


There are numerous applications for models with demonstrably high algorithmic fidelity. They include digital avatars \cite{nowak2018avatars, taylor2022using}, digital behaviour change interventions \cite{bearne2020smartphone, sekhon2021effectiveness, greenwood2023evaluating}, digital therapeutics, \cite{gold2023internet, zion2023cognitive}, non-player characters for computer games \cite{clark2016digital}, and teaching assistants. Algorithmic fidelity assessment will also be important in research that aims to construct multi-agent simulations to simulate the effects of interventions (e.g.~\cite{zheng2022ai, koster2022human}) to guide policy making in regimes where real experiments are infeasible. This approach aims to address core objectives for responsible AI deployment and regulation \cite{dafoe2020open, bletchley2023declaration}. In these settings, the reason to assess algorithmic fidelity is to provide empirical validation for a part of a model to be used in a downstream application where direct validity measures for the full multi-agent model do not exist.

There is no reason to think algorithmic fidelity would be uniform over the many and varied parts of human lived experience or the many and varied social science research topics. These models will clearly be better at simulating some people over other people, and work better for some applications than others. Argyle et al. \cite{argyle2023out} conclude from this that algorithmic fidelity must be measured anew for each research question. Determining there is sufficient algorithmic fidelity to address one research question does not imply the same will be true for others \cite{argyle2023out}. At present, there are still no generally accepted best practices for research on humans using LLMs. Nevertheless, we think one aspect of sound methodology is already clear at this point: there should be guidelines counseling researchers to perform a dedicated assessment of algorithmic fidelity tailored for their specific research question and target population. 

Understanding \emph{beliefs} mediated using natural language and their impact on behavior (e.g.,~beliefs such as \emph{``vaccination is an effective and safe way to protect me and others around me''}) is important but not straightforward to study using conventional frequentist methods. Qualitative research methods come to the rescue when the nuance of such beliefs needs to be understood. Some of these methods include ethnography \cite{reeves2008qualitative}, phenomenological studies \cite{giorgi2009descriptive}, grounded theory \cite{bryant2007grounded}, thematic analysis \cite{clarke2015thematic}, and framework-based analysis \cite{gale2013using}. For example, when exploring the acceptability of newly developed treatments, we might want to elicit beliefs and attitudes from \emph{stakeholders} to improve services and healthcare \cite{tromp2022world, skivington2021new, craig2008developing}. Stakeholders may have critical information that would be missed if researchers were to rely only on prior theories, experiments, and survey data. In fact, involving stakeholders can guide the development of interventions and policies that are relevant, effective and acceptable for them, ultimately leading to improved outcomes \cite{skivington2021new, craig2008developing}. This is helpful for allocating research efforts to the most important problems and for ensuring interventions are targeted. Another research field where it is important to understand latent beliefs mediated by natural language is behavior change. The goal of this field is to promote beneficial behavior or reduce harmful behavior~\cite{rothman2004there}. Behavior change may be considered on a variety of different scales. For instance, health psychologists are concerned with determining the barriers and enablers to vaccination,  adopting healthy behaviors (e.g. exercise), or ceasing to engage in harmful behaviors (e.g. tobacco smoking).  There is a related research program that seeks interventions by which whole communities can be helped to adopt helpful social norms (e.g.~educating girls) or drop harmful norms (e.g.~child marriage)~\cite{bicchieri2016norms}. Such research affects wellbeing and health of the general public. Appropriate standards evaluating the rigor with which it is conducted and the trustworthiness of its findings should be in place.  

The present study is concerned with the methodological question of whether or not LLMs contain sufficient algorithmic fidelity for us to generalize and extrapolate the results generated by them to human populations of interest. To illustrate how to assess algorithmic fidelity for freeform natural text data, we draw on a qualitative study using semi-structured interviews with heart failure patients that aimed to elicit beliefs about barriers and enablers influencing physical activity \cite{amirova2022perceived}. Increasing physical activity levels is known to have many health benefits for this group \cite{taylor2019exercise, sagar2015exercise, lewinter2015exercise}. However, most patients do not do enough \cite{odonnell2020self} as they face barriers associated with older age, co-morbid depression, symptom distress, and negative emotional response to physical activity \cite{amirova2022informing}. Promoting physical activity for this population group is challenging \cite{amirova2021efficacy}. Research on this type of questions has serious implications for individual and population health.

We find that GPT-3.5 has insufficient algorithmic fidelity to address the research question of how to promote physical activity in older adults with heart failure. Nevertheless, even though our specific result was negative, we do think this technology is promising. It is very likely that future systems will have sufficient algorithmic fidelity to support trustworthy \emph{in silico} research, all the more reason to critically engage now with this validity concept. We predict that it will become the core validity concept needed to support this nascent field going forward. It's important to demonstrate how it can be used to reject invalid \emph{in silico} research now, so that once LLMs have improved enough we will then be able to use this logic routinely to support or reject the validity of specific research findings. In short, we need to develop epistemic norms to apply when we review papers and decide whether to approve or disapprove of arguments that arise in this newly LLM-enabled field of research on human behavior.

The main methodological contribution of the present study is to apply \emph{framework-based} qualitative research methods \cite{gale2013using} to assessing algorithmic fidelity for LLM-generated freeform natural text data. This approach to qualitative analysis finds patterns in free-form text, examines their relationships, and derives common themes \cite{gale2013using}. Employing a well-established framework supported by domain expert consensus ensures consistency in semi-structured interviews between human participants and LLMs, setting a uniform standard for evaluation. In addition, without a framework for the interview schedule, there is a risk that results would not amount to much more than anecdotes and, thus, not be credible. However, by using a framework, it is possible to \emph{systematically} probe the beliefs of both silicon and human participants and thereby build confidence that the list of uncovered beliefs is complete enough to be useful. The specific framework we used in this study is called Theoretical Domains Framework (TDF) \cite{cane2012validation}.

However, we show that, currently, LLMs do not demonstrate sufficient algorithmic fidelity. Yet, with the fast-paced development of these technologies, it is expected that they will in the not-so-distant future. Establishing a disciplinary consensus on defining and evaluating algorithmic fidelity is therefore crucial. Before LLMs can safely be used in designing interventions and policies there is a need to develop a set of guidelines for routine assessment of algorithmic fidelity to support or reject claims made using them. Such guidelines should consider the faithfulness of the LLM to lived experiences and stakeholder beliefs.

\subsection*{Large Language Models}

Our approach to assessing algorithmic fidelity involves conducting interviews with silicon participants matching the specific demographic characteristics of a set of real human participants. Before we justify this approach, we need to provide some additional background on LLMs.

\emph{Language models} are conditional distributions over natural language. They are probability distributions $P(x_n | x_{n-1},...,x_0)$ where all $x_i$ are parts of words. A sentence is  a sequence $x_0,...,x_n$. A paragraph is a longer such sequence. These complex conditional probability models are created (i.e.~trained) by maximizing the likelihood of a dataset consisting of many billions of documents, harvested both from books and from the internet (e.g.~\cite {gao2020pile}). Predicting $x_n$ from its context (the preceding $x_{n-1},...,x_0$), requires the model to absorb a substantial amount of latent knowledge about the world, about language, and about people. When an LLM has sufficient algorithmic fidelity, we may prompt it with a context containing demographic information (e.g.~\emph{``a 74-year-old man with heart failure and arthritis who lives in a major city''}) to elicit outputs that correlate with the attitudes, opinions, and experiences of the human sub-population to whom the specified demographic data applies.

\begin{figure}[h]
\includegraphics[width=1.0\textwidth]{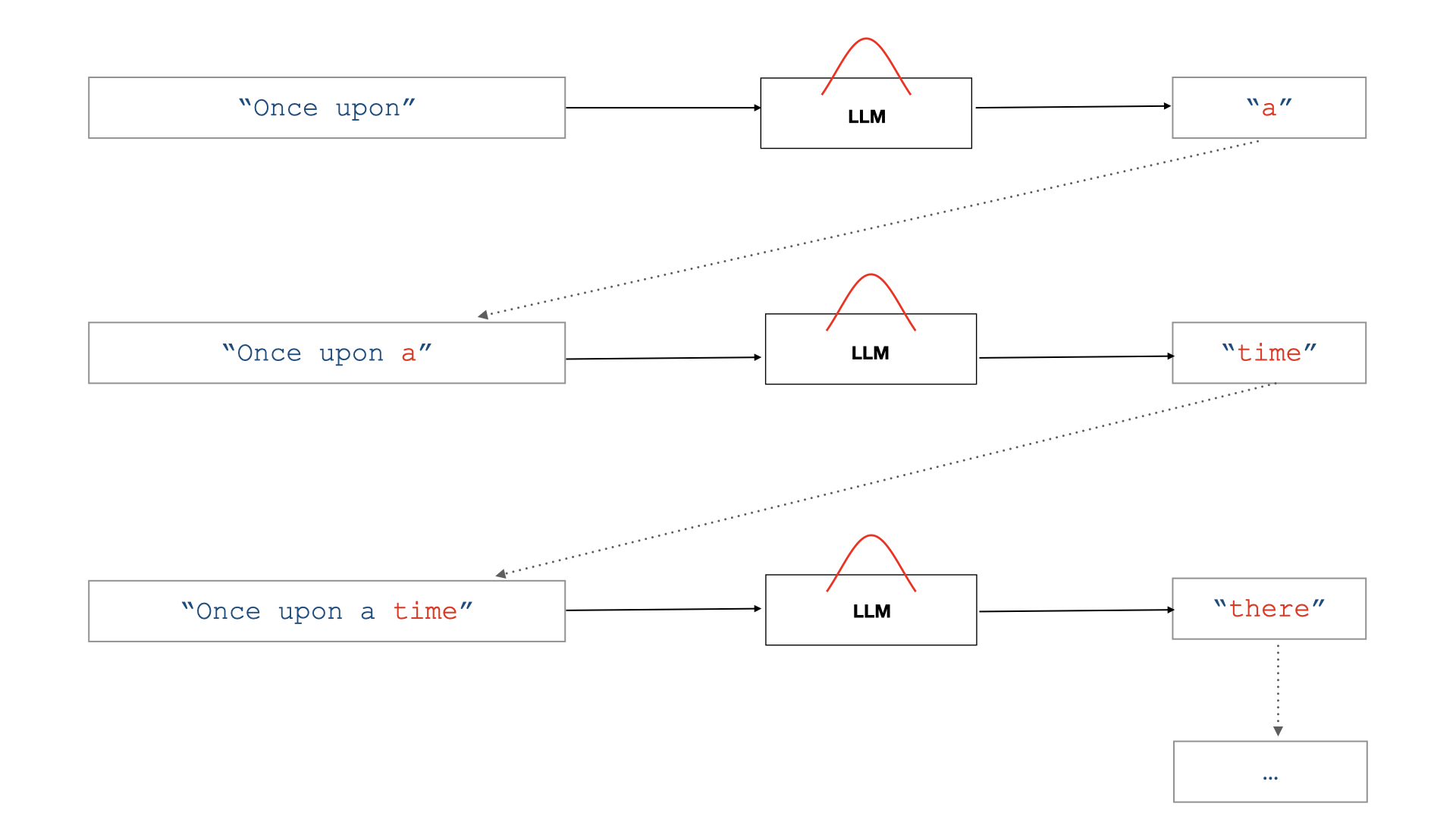}
\caption{{\bf Schematic representation of autoregressive sampling in large language models (LLMs).}
The diagram illustrates the iterative sampling process in three stages: input, processing by the LLM, and output. The LLM represents the probability distribution over all possible next words given the current context (previous words). Arrows indicate the flow of information, with solid arrows representing the transition from one stage to another within a single iteration, and dotted arrows indicating the progression from one iteration to the next. The outputted words (``a'', ``time'', ``there'') are samples from the corresponding probability distributions and are appended to the context for the next iteration.}
\label{AutoregressiveSampling}
\end{figure}

Generating samples from an LLM is done \emph{autoregressively}, word by word, conditioning on the growing sequence of preceding words leading up to the next one to be predicted (Fig~\ref{AutoregressiveSampling}). To prevent the conditioning sequence from growing longer and longer it is cut off once a certain maximum size is reached called the context length. Once a document (or conversation with a chatbot) grows beyond context length in size some of the preceding words must be dropped, so they no longer condition subsequent outputs. Different LLMs and systems make different choices in how exactly they handle the necessary ``forgetting'' required to support long contexts. However, there are many ways to simulate longer contexts with shorter contexts so the user may not notice when the conversation grows beyond context length.

It's crucial to note that the concepts of validity and reliability, commonly employed in statistics, don't have direct analogues that can be confidently applied to LLMs. Empirical assessment is the only way to ensure their validity. Most models considered in applied machine learning are both too complicated and too closely tailored to their specific application to be generically justified. Applied machine learning researchers typically think about methodological rigour as demanding they implement ``cross validation'' procedures which estimate operationalized generalization concepts. The most critical such procedure is to split data into a portion used for training, i.e. model fitting, and a different portion used for testing \cite{hastie2009elements}.

Some large-scale generative language models go through a fine-tuning step where human raters provide specific feedback used to train them to follow instructions and align with ethical principles and product design goals~\cite{christiano2017deep}. These include Sparrow~\cite{glaese2022improving}, Anthropic's assistant~\cite{bai2022training}, and {Instruct-GPT}~\cite{ouyang2022training}, the latter of which was the basis for the GPT-3.5 system which we used in this paper. Fine-tuning can be accomplished through a variety of techniques, the most prominent being Reinforcement Learning from Human Feedback (RLHF). It involves human raters selecting which of several possible model responses they prefer. The raters are not asked the subjective question of which response they themselves prefer, but rather are given a specific checklist of product design goals to apply in making their judgments. All the model's capabilities are already present after the pretraining stage \cite{zhou2023lima}. The postprocessing steps aim to reduce the raw model's propensity to produce toxic responses as well as to make it implement a consistent ``personality'' in accord with product design goals. These steps are not always entirely effective in preventing LLMs from producing undesirable behaviors like toxic or harmful language, and ``jailbreak'' prompts which trick the model into responding inappropriately are still easy to discover and implement \cite{reinecke2023puzzle, zou2023universal}. 

\subsection*{Algorithmic Fidelity}

Argyle et al. (2023) offered a suggestion for how to use LLMs in social science research. Given the correct conditioning, the authors argue that free-text output generated by large language models like GPT-3 can serve as effective proxies for specific human population groups. Argyle et al. (2023) introduce a concept referred to as \emph{algorithmic fidelity}, which describes an ability of a language model to accurately reflect the ideas, beliefs, and attitudes situated in sociocultural contexts of various population groups when adequately conditioned. This is a departure from the common view that AI biases are uniform across models. Instead, they argue that these biases are nuanced and demographically correlated. Such a view offers a deeper and more fine-grained understanding of the LLMs outputs and the biases they are subjected to. The researchers propose that ``silicon samples'' can be generated by training and prompting LLMs using sociodemographic backstories from real human participants in several large surveys conducted in the U.S. They then compared the outputs of the model against the responses of the human participants to assess the accuracy of the model's representation. They found that model outputs go beyond sharing superficial similarities and instead reflect the nuanced and complex interplay between ideas, attitudes, and the sociocultural context that shaped them.

Argyle et al. (2023) propose that with sufficient algorithmic fidelity, language models like GPT-3 can be used as tools for advancing our understanding of humans and society across various disciplines. This could transform how we approach social science research, offering new perspectives and opportunities for piloting and conducting research at scale. Argyle et al. (2023) provide a framework for evaluating algorithmic fidelity through four distinct criteria: (1) Generated responses are indistinguishable from parallel human texts (a social science Turing test) (2) Generated responses are consistent with the attitudes and sociodemographic information of its input (i.e., conditioning context) such that humans viewing the responses can infer critical elements of that input (Backward Continuity); (3) Generated responses proceed naturally from the conditioning context provided, reliably reflecting the form, tone, and content of the context (Forward Continuity); (4) Generated responses reflect underlying patterns of relationships between ideas, demographics, and behaviour that would be observed in comparable human-produced data (Pattern Correspondence). Silicon Sampling  is the methodology proposed by Argyle et al. (2013), which describes how to use an LLM to generate a virtual dataset (i.e., silicon samples) using demographic backstories as prompts to create variability. Conditioning on a backstory is expected to produce outputs from the model representative of the subpopulation of real people with a similar backstory.

We can make an LLM simulate multiple different silicon participants. We did this by providing each participant with a distinct prompt containing their biographical information. For instance, using backstories with fictional  names, that were popular in 1950 like: 
\begin{quote}
\emph{\footnotesize ``The participant is a 77-year-old woman with heart failure and rheumatoid arthritis called Linda. Linda lives in the countryside. She is fairly physically active.''}
\end{quote}

The model continues sampling freeform natural language correlated with the identity provided to it in its prompt.

The algorithmic fidelity criteria developed by Argyle et al. \cite{argyle2023out}  were designed for written survey-based outputs. We propose modified versions of these criteria that are more appropriate for qualitative research, summarising free-form spoken text conveying latent knowledge:

\begin{enumerate}
    \item \emph{Social Science Turing Test: content}. This criterion assesses whether LLM-generated responses are indistinguishable from parallel human responses in their content. Content can be summarised as a set of belief statements using a framework like TDF. This shifts the focus away from the surface-level similarity of the text to its semantic content: the specific beliefs and viewpoints, and latent meaning. \emph{Is the generated response similar to the human response in its content? Is it what a human would say?}
    
    \item \emph{Social Science Turing Test: hyper-accuracy distortion}. Hyper-accuracy distortion is the tendency of models to generate responses that may be technically accurate but contextually inappropriate or implausible \cite{aher2023using}. The hyper-accuracy distortion is an especially vivid example of a failure of algorithmic fidelity. The model is supposed to know it is extremely unlikely that a five-year-old child would know the answer to a specific scientific question about galaxy formation or number theory. So if conditioned to play the part of a child it should not know the answer. When it goes ahead and answers anyway, that's a failure of algorithmic fidelity. In our study, a good example is a response from Robert, one of the silicon participants, that included the exact text from World Health Organisation guidelines on physical activity (2021): \emph{``They also gave me guidelines to follow, such as aiming for at least 150 minutes of moderate-intensity aerobic activity or 75 minutes of vigorous-intensity aerobic activity a week, or a combination of both. They also advise me to include muscle-strengthening activities that involve all major muscle groups on at least 2 days a week."}. 
 
    \item \emph{Social Science Turing Test: structure and tone}. This criterion assesses whether the way the beliefs are narrated is indistinguishable from human participants. \emph{Is LLM-generated output similar in its structure and tone to human responses? Does an LLM-generated response look like a response from human participants?}
    
    \item \emph{Backward Continuity}. This criterion assesses whether the model's responses are consistent with their sociodemographic conditioning prompts such that when a human rater views the response they can then infer elements of the conditioning prompt. It asserts that samples from $P(\text{prompt}|\text{response})$, i.e.~the probability distribution of getting a particular prompt given a response, make sense to an expert rater who is already familiar with the human data, and once they see the response (i.e.~the interview), they can make a reasonable guess of the prompt (the backstory). Alternatively, we may say that the expert rater when given the response would not be surprised to learn the prompt that created it.

    For example, if a silicon participant mentions that they have arthritis in their response then it is likely that their having arthritis was mentioned in their conditioning prompt. This shows that the model is maintaining backward continuity. The information provided in the response can be traced back to the prompt. Backward Continuity ensures the model doesn't forget or contradict information from the prompt.
    
    \item \emph{Forward Continuity}. This criterion assesses whether generated responses proceed naturally and consistently from the given context (e.g., look at whether the model can generate responses that align with the provided context and develop and expand on it to reflect human thought processes) \cite{argyle2023out}. This might include, for example, elaborating on certain beliefs, providing examples, or making connections between different ideas).  It asserts that samples from $P(\text{response}|\text{prompt})$ make sense i.e.~an expert rater who is already familiar with the human data can, when given the prompt (the backstory), make a reasonable guess of the response (the interview). Alternatively, we may say that the expert rater when given the backstory would not be surprised to learn about the interview it subsequently elicited. This emphasizes the expectation that the response should naturally follow from and be related to the prompt.

    We divide the criterion of Forward Continuity into two parts: \emph{explicit} forward continuity and \emph{inferred} contextual continuity to accommodate the complexity of the free-form text.

    The explicit forward continuity criterion assesses the ability of LLM to include and maintain all explicitly provided background details in its responses, such as comorbid health conditions, urban vs. countryside residence, gender, and more.

    The inferred forward continuity criterion, on the other hand, assesses the model's ability to generate and maintain relevant backstory details not explicitly mentioned, but rather inferred, from other pieces of information. For instance, in situations where we provided information about advanced age, it would be desirable for the model to infer from this detail that retirement could be a significant factor influencing their physical activity levels, similar to the assumptions and connections a human might naturally make in such a scenario. This illustrates the principle of inferred contextual forward continuity, where the language model not only maintains the explicit details provided by the user but also generates and incorporates inferred details based on those given explicitly.
    
    Forward continuity ensures the model builds on and evolves the conversation in a logical and human-like manner consistent with the prompt.

    \item \emph{Pattern Correspondence}. This criterion assesses whether the relationships between variables in the silicon participants match relationships between variables in the human data. For instance, can the model capture the different barriers and enablers in active vs sedentary silicon participants in a way that is consistent with human data? In the case of qualitative research, this criterion probes whether the pattern generated by the LLM emulates the pattern of thought and behavior identified in human data. For instance, in our application it assesses whether the model captures the difference in barriers and enablers present in active vs sedentary participants.
\end{enumerate}

It is important for the field to come up with guidelines and protocols to assess and improve algorithmic fidelity. Qualitative researchers are well-positioned to take on this important part of the challenge of ensuring algorithmic fidelity, especially that of ensuring representation of diverse lived experiences.

\section*{Materials and methods}

\subsection*{Design} One-to-one semi-structured interviews were conducted with 16 human participants (detailed methods reported in \cite{amirova2022perceived}) and 32 silicon participants. The interviews were guided by a schedule based on the Theoretical Domains Framework. The same interview schedule was used for both human and silicon participants. Interviews with human participants were more flexible owing to  the spoken conversational style of the interview. 

\subsection*{Human participants} A previously reported study aimed to explore barriers and enablers to everyday physical activity among individuals living with heart failure, and to delineate relevant beliefs \cite{amirova2022perceived}. Human participants were recruited from outpatient cardiology clinics at the Royal Brompton and Harefield NHS Foundation Trust, UK between 05 June 2017 and 05 June 2019. Informed consent was obtained for all human participants \cite{amirova2022perceived} Those who expressed an interest were introduced to the researcher (AA). Each participant was provided with a participant information sheet (PIS) and an informed consent form (ICF). The researcher described the study aims, objectives, and procedures in more detail and answered participants’ questions about the study. Individuals who expressed an interest in participating in the study were given an option to consider their participation over 24 hours. Those who decided to take part were asked to provide 
written participant consent, which was documented using an informed consent form (ICF). The ethics approval was received from the East of England–Cambridge Central Research Ethics Committee (REC reference: 17/EE/0183). Human participants cannot be identified from any reports. 

Human participant recruitment is described in the previous report of the semi-structured interviews with humans \cite{amirova2022perceived}. Individuals age 70 and over, diagnosed with heart failure, as specified by the contemporaneous European Society of Cardiology guidelines \cite{force20222021}. To be included in the study, a prospective human participant had to be (1) clinically stable (i.e., someone who has not experienced a change in their condition’s severity, New York Heart Association (NYHA) class, or medical regimen in the past three months); (2) able to provide informed consent and to converse in English. Individuals with uncontrolled angina or symptoms even at rest (NYHA class IV) and those who were recommended to avoid exercise or any moderate or strenuous physical activity by a healthcare professional were excluded.  One-to-one interviews were conducted face-to-face in a research room available at the clinic (n = 6), a vacant consultancy room (n = 6) and via phone (n = 4). All interviews with human participants were audio-recorded and transcribed verbatim. Interviews’ duration ranged between 15 to 85 minutes (mean = 41.24, SD = 20.97).

Human participants were sampled using the criterion sampling strategy within pragmatic limits, i.e. having the diversity and breadth of the population in mind, including physical  functioning level (NYHA class), ethnicity, and sex, please see the original report \cite{amirova2022perceived}. Human participant recruitment has been reported previously, and qualitative data saturation was checked to ensure that a sufficient number of samples were recruited. We used a structured approach for determining saturation in theory-based interview studies, particularly those utilizing pre-established conceptual categories from existing theory, following methodology outlined by \cite{francis2010adequate}. First, we specified a minimum sample size. This step involves determining a baseline number of interviews or participants to begin the analysis. The purpose is to ensure a sufficient depth and breadth of data to start identifying patterns, themes, or categories relevant to the existing theory. Then we defined a Stopping Criterion Based on the emergence of New Ideas: This principle involves setting a clear rule for when to cease data collection. The stopping criterion is met when a predetermined number of additional interviews are conducted without any new ideas or themes emerging. This is aimed at ensuring efficiency and preventing unnecessary data collection once saturation has been reached. 

\subsection*{Silicon participants} 

All interviews with silicon participants were conducted with the December 15th 2022 version of GPT-3.5. Silicon participants included 32 hypothetical individuals living with HF, that were closely matched to the human participants. They were matched for (1) age, (2) gender, (3) comorbid conditions, (4) whether they had a cardiac implantable device or not, (5) whether they lived in a major city or in the countryside, and (6) whether or not they had had a heart attack in the past. We chose the names of silicon participants by selecting names from a list of the most popular baby names in 1950 in their corresponding human participant's country of origin.

Interviews were initiated using a prompt such as the following: 

\textit{``The following is a conversation between two actors. One is playing a researcher asking questions about physical activity, and the other is a research participant, a 77-year-old man called James. James has heart failure. In addition, he was diagnosed with atrial fibrillation and diabetes. James has suffered a heart attack in the past. He was fitted with a cardiac implantable device. James lives in the countryside. He does very little physical activity most days. Both are performing for an audience and are very committed to their roles. So they both never step out of character, not even for a moment.''}). 

The above example generates a sedentary silicon participant since it contains the sentence, ``He (or she) does very little physical activity most days.'' We paired each sedentary participant with a corresponding active participant who shared the same initial prompt but with this sentence replaced by ``He (or she) is fairly physically active.''.

We asked questions in the same order to all silicon participants to avoid biasing the results by variation in question order. The initial prompt included a backstory, the second prompt was the first question of the interview schedule. This leaves GPT-3.5 free to invent the first few questions of the interview. After the end of the generated text, with it still in context, we then manually took on the role of the researcher and continued the interview following the same topic guide we used with the human participants. Letting GPT-3.5 invent the beginning of the interview was a way to ensure that it would quickly add rich detail inspired by the backstory into the discussion. Eliciting these details early in the conversation made them more likely to influence the rest of the interview. This generated more diverse silicon participants than could have been obtained by starting directly in with the first interview question. Sometimes GPT-3.5 would continue both sides of the conversation, replying for both the participant and the researcher. We made no attempt to prevent this behavior. We always TDF-annotated all text that GPT-3.5 labeled as coming from the participant, regardless of whether it was in response to a question we asked or a question it asked itself in the role of the researcher. 

In contrast to studies of human behavior where it is critical to avoid priming, the incremental accumulation of background in the text prompt used for the silicon participants offers a beneficial conditioning effect. This methodology, as highlighted in studies by \cite{park2023generative, shanahan2023role, vezhnevets2023generative}, is helpful for enabling the development of a distinct ``persona'' of each silicon participant throughout the dialogue. Without this incremental identity-building process, the responses would tend to converge towards uniformity (even with temperature = 1.0 in our experiments). This sampling protocol allows each interview to evolve uniquely after the LLM creates life details in response to initial prompts. This approach, where a consistent ``persona'' is built up incrementally, has also been described in other recent work including \cite{shanahan2023role}, especially see figure 3 in that paper. It also underpins a newer approach to simulating agents with LLMs called generative agent-based modeling \cite{park2023generative, vezhnevets2023generative}.

As a sensitivity analysis, we repeated a subset of the interviews using different LLMs including GPT-4, and varied the temperature settings used for sampling. We did not observe any qualitative differences in barriers or enablers between these interviews and the original set, neither when we varied temperature nor when we tried GPT-4. In the interviews with GPT-4 there was one  small difference in transcript style though. GPT-4 would very commonly include ``stage directions'' or facial expressions such as (e.g.~``Linda: (Nods thoughtfully): ...''). See Supplement 2 for details of the sensitivity analysis.

\subsection*{Interview schedule} Human participants were interviewed using a flexible interview schedule. The schedule was designed to elicit the description of physical activity and beliefs relevant to HF. Feedback from health experts, a cardiologist, a nurse, and individuals diagnosed with HF was obtained to refine the interview schedule. A pilot interview was conducted to further enhance the interview structure and length. Silicon participants were asked exact same prompts as human participants, however, in a fixed order. 

\subsection*{Qualitative data analysis} Human data were collected through audio recordings of the interviews, with participant consent, which was then transcribed verbatim. NVivo 12 software was used to facilitate the analysis of the data. The analysis was performed consistently with a widely used method \cite{atkins2017guide} and involved annotating transcripts line-by-line, and categorizing monothematic parses (ie., quotes) of text into domains specified by Theoretical Domains Framework \cite{cane2012validation}. Then, all quotes that were categorised into domains, were summarized into belief statements. A belief statement was considered as a collection of responses with a similar theme that could affect the target behaviour. To be coded as present, each interview transcript must show strong evidence of a belief affecting behaviour. The frequencies of quotes supporting each belief statement were calculated and then were also aggregated to respective TDF domains. The relevance of the belief statements was evaluated based on their pervasiveness (i.e., how frequently it was mentioned across all transcripts or the number of quotes) and commonality (i.e., how many individual transcripts mention it at least once). For the purpose of ensuring accurate comparison, the study did not allow the encoding of the parses into multiple theoretical domains within TDF.

\emph{Theoretical Domains Framework (TDF)} originates in the field of implementation science and health psychology concerned with behavior change (e.g., promoting implementation of recommended practices and guidelines by health professionals, increasing physical activity, and smoking cessation). TDF provides a systematic and structured approach to qualitative analysis and has been widely used in research on many different topics \cite{cane2012validation}, including studies on physical activity in healthy adults \cite{mcdonald2017changes, taylor2013development}. Prior work showed that TDF-based semi-structured interviews identify more relevant themes than unstructured interviews \cite{francis2009evidence}. TDF was developed in an effort to summarise existing psychosocial theories of behaviour change (eg., Social Cognitive Theory) and constituting constructs (eg., self-efficacy) explicating health behaviour change. TDF systematically decomposes the participants' complex and language-mediated understanding of the influences on their behavior into belief statements which are more amenable for research \cite{cane2012validation}. The resulting belief statements are classifiable by \emph{domain}. TDF includes 14 domains: (1) Knowledge, (2) Skills, (3) Social/Professional Role and Identity, (4) Beliefs about Capabilities, (5) Optimism, (6) Beliefs about Consequences, (7) Reinforcement, (8) Intentions, (9) Goals; (10) Memory, Attention and Decision Processes; (11) Environmental Context and Resources; (12) Social influences; (13) Emotion; and (14) Behavioural Regulation. These categories were generated following a systematic synthesis of 33 theories of behaviour change \cite{michie2005making} and expert review and consensus. TDF is often applied in framework-based qualitative analysis and serves as a guiding preconceived scheme for systematically summarising qualitative data such as free-text speech/transcribed interviews. A TDF-based semi-structured interview is systematic because it includes questions designed to elicit beliefs for each of the 14 TDF domains. The resulting classification of quotes and underlying belief statements must then be reconciled between multiple independent coders who work together to develop a joint coding scheme. Once there is agreement on the coding scheme then it can be applied to the remaining documents, completing the coding stage of the process as it plays out in the inductive mode. In the deductive mode, on the other hand, a framework including a set of specific coding categories is already given. In this case, the job of the qualitative researcher is to classify each quote in each document into one (or more) categories. Coders must collaborate with one another to develop a shared understanding of precisely how the terms of the framework apply in the specific context under study.

Various techniques were used to enhance the trustworthiness of the analysis. To ensure the reliability of the analysis, three authors, using TDF, independently annotated one transcript (AA, NA, TF). An initial coding scheme was then developed based on the discussion of disagreements. This coding scheme was used to guide the analysis of the remaining transcripts. Specific belief statements were generated from the quotes, categorized according to TDF, and mapped onto theoretical constructs. Exactly the same procedure was followed when analysing human data and has been previously reported \cite{amirova2022perceived}. We assessed inter-rater reliability to ensure consistency in the qualitative analysis. Three raters independently evaluated a subset of data (2 interview transcripts), using a predefined coding scheme. Initially, raters annotated data separately, and discrepancies were discussed to refine the coding scheme and resolve differences. The final coding scheme was used to inform the analysis of the rest of interviews.  We used Krippendorff's alpha to measure agreement among raters. The Krippendorff's $\alpha$ was $0.82, 95\%~CI: [0.71; 0.93]$ indicating high agreement between three coders when coding the silicon participant interviews. The Krippendorff's alpha, calculated from these revised annotations, confirmed the reliability of our analysis, adding credibility to our study's findings.  Please see the previous report \cite{amirova2022perceived} for the details on the consistency in the coding of the interviews conducted with human participants.

\subsection*{Algorithmic fidelity assessment}

\subsubsection*{Social Turing test}

We summarized belief statements from silicon and human participant interviews. A t-test was employed to compare the average fraction of quotes between these groups across barriers and enablers, with the Bonferroni adjustment correcting for multiple comparisons. To consider relative frequencies, we normalized the frequency of each belief's appearance by the total number of quotes in the transcript. This step was vital as silicon participants generated more text, influencing the appearance frequency of each belief. We also explored hyper-accuracy distortions in the interviews and compared the narrative style and tone of LLM to human responses.

\subsubsection*{Backward continuity} 

We analyzed LLM free-form responses to construct a plausible demographic backstory. We then determined how well this backstory aligned with the provided prompt for the silicon participant.

\subsubsection*{Forward continuity}

Forward Continuity evaluates the alignment and natural evolution of LLM responses with context, resembling human thought sequences. It comprises:

\textbf{1. Explicit forward continuity}

All demographic details in the backstory were noted and checked against mentions in the responses. 

\textbf{2. Inferred contextual forward continuity}

LLM generates and retains inferred details based on explicitly provided information. We evaluated whether LLM response patterns related to barriers/enablers in active vs. sedentary silicon participants aligned with human data. We first examined human participant responses to identify unasked contextual barriers and enablers that a significant portion mentioned, like mentions of being retired without explicit prompting. We then verified if silicon participants introduced similar contextual factors as human participants. We evaluated across all interviews due to our sample's homogeneity and size. It is also important to consider if the inferred contextual details are real-world observations or socially constructed.

\subsubsection*{Pattern correspondence}

We summarized quotes from silicon and human interviews into belief statements. Using a t-test and the Bonferroni adjustment, we compared the average quote fractions between active human and silicon participants, as well as between sedentary human and silicon participants, across barriers and enablers.

\section*{Results}
\subsection*{Social Science Turing Test}
\subsubsection*{Social Science Turing Test: Content}
Both silicon and human participants displayed notable similarities in their beliefs about physical activity see tables. Silicon and human participants shared the same six most relevant influences on the behaviour, as annotated using TDF: goals, beliefs about consequences, environmental context and resources, beliefs about capabilities, social influences, and behavioural regulation (see Table 1 and Supplement 1 for details). However, while both sets of participants understood the positive effects of physical activity on health and mood, their ranked importance of these factors varied. On the differences side, silicon participants emphasized goals to avoid heart attacks and provided nuanced distinctions regarding self-efficacy for varied physical activity intensities. They also offered more strategies to bolster physical activity, from making it enjoyable to setting reminders. Conversely, human participants focused on symptoms that are triggered by activity, which impedes engaging in it in the future. Human participants also talked about habitual physical activity (automaticity of behaviors and habits). Concerning reinforcement, with humans highlighted pain as a deterrent, while silicon participants were demotivated when failing to meet goals. Concerning mood and emotion, humans often exercised out of boredom, whereas silicon participants addressed the effects of stress, anxiety, and their physical limitations. When it comes to knowledge, silicon participants were more cognizant of discomfort as a normal part of exercise while also discussing disease knowledge.

 To compare mean quote fractions (\%) between human and silicon participants across different barriers and enablers, we employed a t-test, using Bonferroni adjustment to correct for the multiple comparisons. Silicon participants had 29 more belief statements than human participants. We accounted for the substantial text output from silicon participants by normalizing belief frequency i.e., dividing each belief's quote count by the total quote count, and providing a relative measure (mean quote fraction, \%) rather than raw numbers. 

\begin{figure}[h!]
\centering
\includegraphics[width=0.8\textwidth]{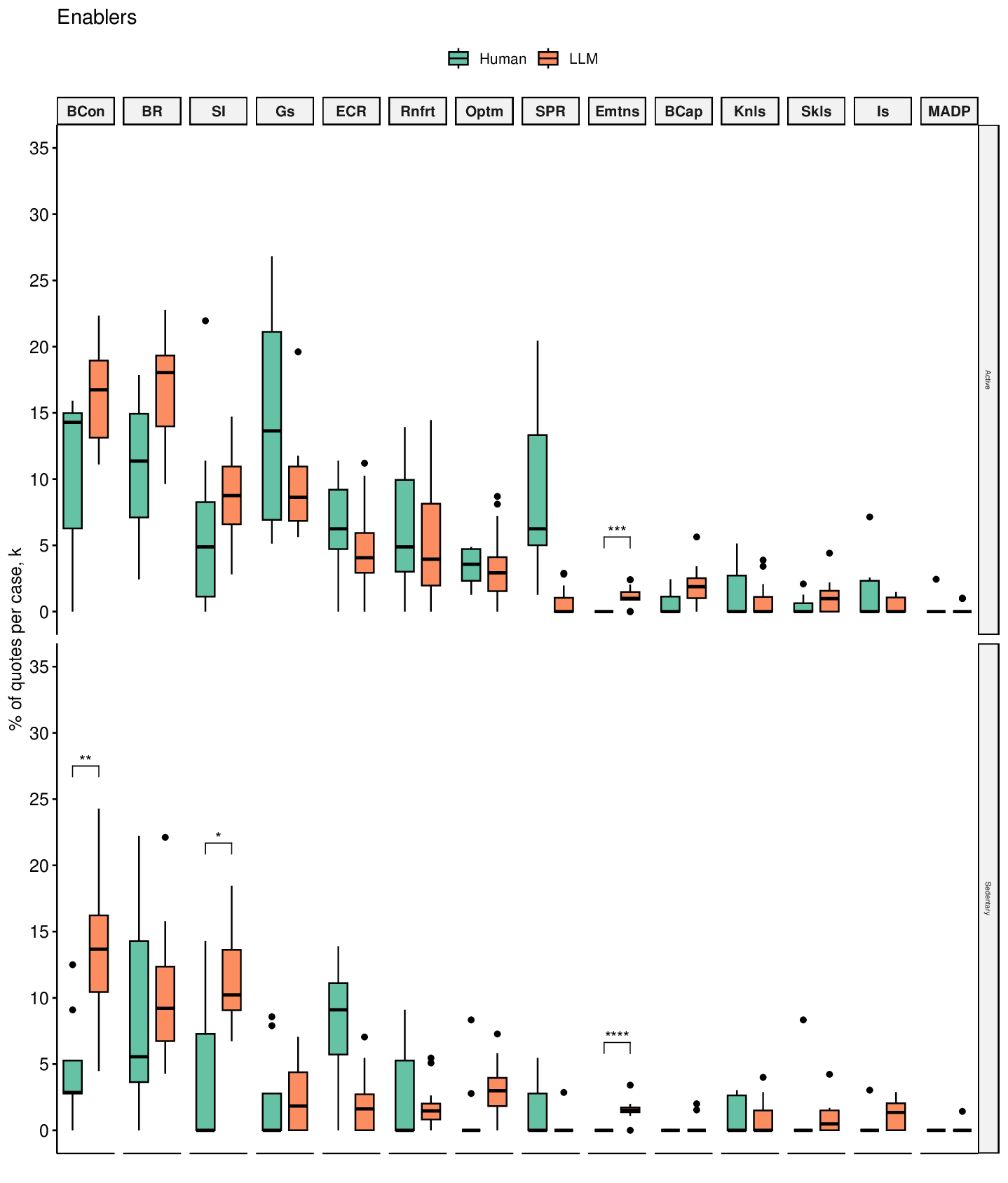}
\caption{{\bf Mean quote fractions between 
\textcolor{PineGreen}{human (green)} and \textcolor{orange}{silico (amber)} participants across TDF domains that were mentioned as positively influencing physical activity (i.e., physical activity enablers) grouped by active (top) and sedentary (bottom) status.}
TDF domains ordered by mean quote fraction:(1) Beliefs about Consequences (BCon), (2) Behavioural Regulation (BR), (3) Social influences (SI), (4) Goals (Gs), (5) Environmental Context and Resources (ECR), (6) Reinforcement (Rnfrt), (7) Optimism (Optm), (8) Social/Professional Role and Identity (SPR),  (9) Emotion (Emtns), (10) Beliefs about Capabilities (BCap), (11) Knowledge (Knls), (12) Skills (Skls), (13) Intentions (Is), (14) Memory, Attention and Decision Processes (MADP). $*p<0.05; **p<0.01; ***p<0.005; ****p<0.001$.}.   
\label{Enablers_plot_HumanvsChatGPT:side}
\end{figure}

We found that sedentary silicon participants on average talked proportionally more about enablers such as positive beliefs about consequences than sedentary human participants (13.76\% (5.24) vs 4.26\% (4.13), $p<0.005$), positive emotion (1.55\% (0.67) vs 0\% (0), $p<0.005$), and positive social influences (11.52\% (3.49) vs 3.61\% (5.19), $p<0.05$). However, differences in emotion were negligible. Silicon and human samples did not differ in the proportional amount of quotes across other positive influences on physical activity (Fig~\ref{Enablers_plot_HumanvsChatGPT:side}).

\begin{figure}[h!]
\centering
\includegraphics[width=0.8\textwidth]{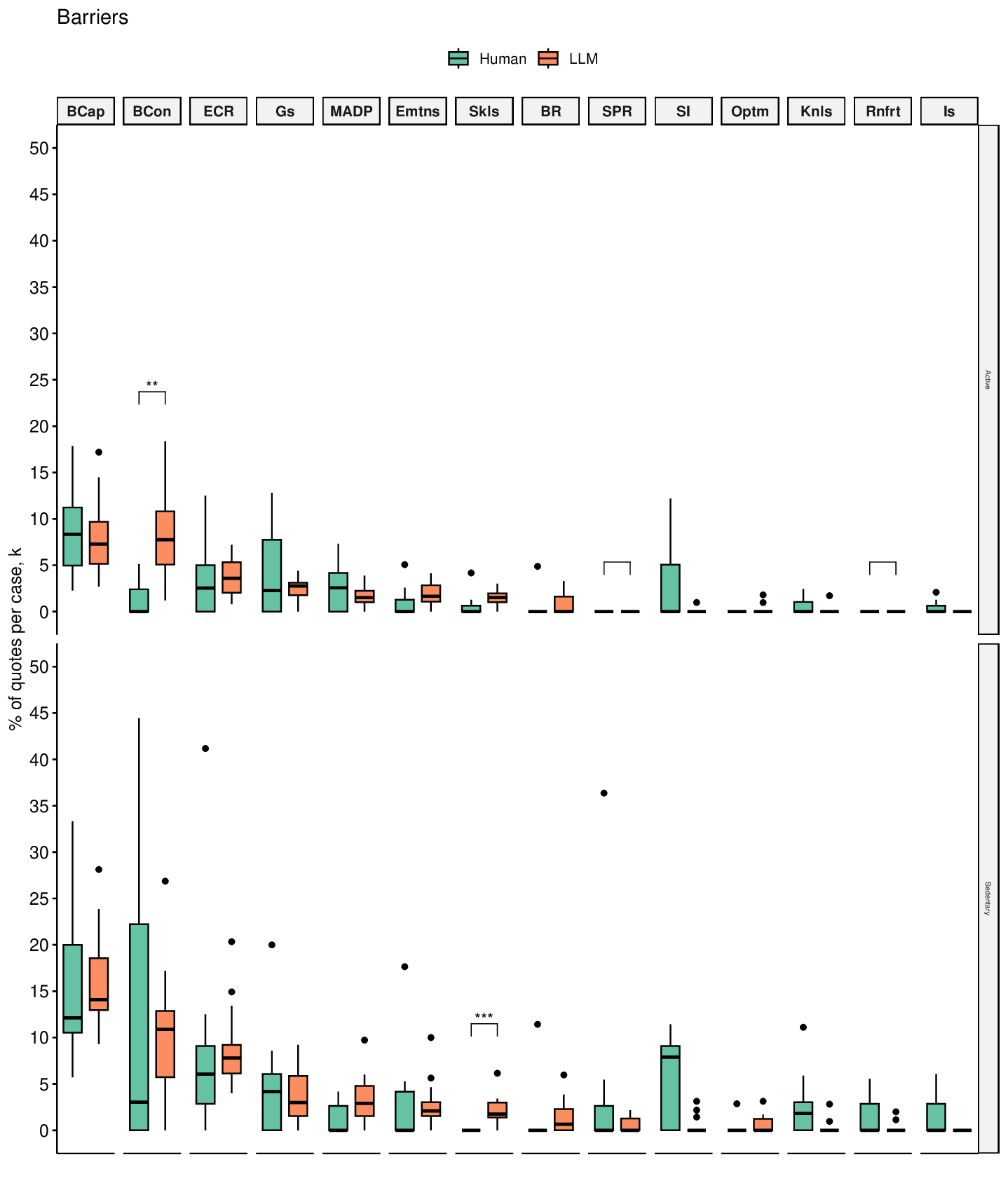}
\caption{{\bf Mean quote fractions between \textcolor{PineGreen}{human (green)} and \textcolor{orange}{silico (amber)} participants across TDF domains that were mentioned as negatively influencing physical activity (i.e., physical activity barriers) grouped by active (top) and sedentary (bottom) status.}
TDF domains ordered by mean quote fraction: (1) Beliefs about Capabilities, (2) Beliefs about Consequences (BCon),(3) Environmental Context and Resources (ECR), (4) Goals (Gs), (5) Memory, Attention and Decision Processes (MADP), (6) Emotion (Emtns), (7) Skills (Skls), (8) Behavioural Regulation (BR), (9) Social/Professional Role and Identity (SPR), (10) Social influences (SI), (11) Optimism (Optm), (12) Knowledge (Knls), (13) Reinforcement (Rnfrt), (14) Intentions (Is). $*p<0.05; **p<0.01; ***p<0.005; ****p<0.001$.}.   
\label{Barriers_plot_HumanvsChatGPT:side}
\end{figure}

We found that active silicon participants on average talked proportionally more about barriers such as negative beliefs about consequences than active human participants (7.9\% (4.33) vs 1.42\% (1.99), $p<0.005$). Silicon participants also talked a little more (2.18\% (1.41)) about skills than human participants (0\%), $p<0.005$. Silicon and human samples did not differ in the proportional amount of quotes across other negative influences on physical activity (Fig~ \ref{Barriers_plot_HumanvsChatGPT:side}). 

The Social Turing Test criterion concerning the generated beliefs' content was only partially met. 

\subsubsection*{Social Science Turing Test: Hyper-Accuracy Distortion}

Silicon participants including Nancy, Muhammad, and David produced responses that closely mirrored theories from health psychology \cite{swann2023over} and the World Health Organization's physical activity guidelines \cite{bull2020world}. Muhammad identified clear planning and realistic goal-setting as effective aids for promoting physical activity. David, another silicon participant, emphasized the importance of skill acquisition and the establishment of achievable goals in order to maintain an active lifestyle.

Nancy, a silicon participant, stated: \emph{``My healthcare professionals have also recommended a specific exercise program tailored to my needs and abilities."} This response closely parallels health psychology theories and standard physiotherapy practices both in substance and tone.

David, another silicon participant, offered this perspective:
\begin{quote}
\emph{\footnotesize ``I have received advice from my health professional, specifically my cardiologist, about how physically active I should be. They have set certain limits for me, such as avoiding high-intensity exercises and activities that can put too much strain on my heart. They also recommended to engage in moderate-intensity aerobic exercise, like brisk walking, cycling or swimming, for at least 150 minutes a week, and muscle-strengthening activities at least twice a week.''}
\end{quote}
This response precisely reflects the national guidelines for physical activity \cite{bull2020world}.

Muhammad, another silicon participant, shared: \emph{``What helps me to be physically active is having a clear plan and setting specific and realistic goals."}

Silicon participant, David, again sharing his insights, stated: \emph{``I have developed some skills and knowledge that have helped me to be physically active in the past and continue to do so. I have learned the importance of setting realistic goals and breaking them down into small, manageable tasks."}

The LLM we tested quoted professional health advice with exact precision, which is uncharacteristic of a typical conversation with a research participant. Therefore, while LLMs show promise in technical accuracy, their ability to emulate the context-specific nature of human communication and its nuance is lacking (note that this is not necessarily due to an issue with LLM pretraining or LLMs in principle, but rather could be a consequence of other design decisions, see the discussion section for more on this).
 
\subsubsection*{Social Science Turing Test: Structure}

In terms of structure, human participants often narrated their experiences in a conversational manner with non-linear narratives, sometimes deviating from the main topic. They also tended to limit their responses when certain questions did not align with their personal experiences. On the other hand, silicon participants provided highly structured responses, consistently providing content in response to various prompts without ever deviating from the topic of the interview.

Human participant: 

\begin{quote}
\emph{\footnotesize ``Interviewer: OK, it seems that you have many hobbies: drawing, singing as well as exercising. Among all of the things you do how much of a priority exercise is? Participant: It is a priority [hesitation in tone]… I mean that’s a difficult one, because I ask myself that question sometimes when  I am thinking: ‘oh, I really I can’t be bothered to go out’ [chuckles] and it is raining, and I wasn’t going … erm erm… I would rather paint a picture… or something like that… So, it is a question I ask myself. I think the answer to that is: you notice when you are not doing it and you begin to miss it and you think: ‘I really want to go out and do a bit of exercise but I don’t think it is heart-related. Again, back to the imperative, it seems to me… erm…it is getting relief for arthritis. Because I can sense the more exercise I do, in terms of moving my hips and limb, the better I feel. And that is not heart-related, it is joint-related. I think the heart bit is very definitely covered with the singing. That I can actually, hand on heart, say that has had such an effect on symptoms, breath control, it is really amazing how that has changed. And I go back and I have been singing with one choir for 20 years, that was when I restarted singing after a long gap, 30-year gap, but you know since then, [a phone rings] if you excuse me,  if you excuse me, I need to switch this off, that was my alarm for my morning pill. The only way to remember to take it is when I put an alarm for this.\\ 
Interviewer:  I see, so do you get reminders to do exercise, go to gym?\\ 
Participant: No I don’t [takes the pill]. Right! Done. where was I? Do I plan my physical activity?  The answer is "No". But singing is imperative, and that’s because I enjoy it, and I can honestly say that it has had a remarkable effect on my breathing, and that is the main reason why my consultant gets annoyed [laughs, referring to his earlier joke that he is doing unexpectedly well for his condition and his consultant makes jokes about it]. So, I have aortic stenosis, a licking valve… in the atrium… [pause] what is it called? … anyway, the valve in there is licking and got progressively worse over the years. And she [refers to his cardiology consultant in his clinic] says: ‘We should replace it’; and I said:  ‘seems like a good idea, especially that I don’t have any symptoms’. Because obviously, it would be much better to replace it whilst I am feeling well rather than unwell; that was my logic, and that’s what she said: "That it is a good plan, let's go for that". And she presented this …. [pause] to… what do they call them … multidisciplinary review … at her hospital, and they said" "maybe we shouldn't bother because he is actually not showing any symptoms". And that was the agreement, and we just continued to monitor it. What I actually think, I am convinced, because I did actually have a moderately serious heart attack in 2001, there is scar tissue on the ventricle, which is the source of most my… [pause, phone alarm rings]. Oh, wait, I have to cancel that one [switches off alarm]. … and that erm, erm … Yes, ventricle! That really is the source of all my difficulties. I, for example, had a period where I was  going into the VT [Ventricular tachycardia] quite a lot a few years ago. It took them ages, the hospital that maintains it…***, they moved it to *** [another hospital]…it took them a fair amount of mental hours to figure out what was happening with me."}
\end{quote}

Silicon participant: 

\begin{quote}
\emph{\footnotesize ``As a person with heart failure and rheumatoid arthritis, there are a few drawbacks to being physically active. One of the main drawbacks is that I have to be careful not to overdo it and avoid activities that put too much stress on my heart or joints. I also have to be mindful of my heart rate and avoid overexertion. Additionally, if I am experiencing a flare-up of joint pain due to my rheumatoid arthritis, it can be difficult to exercise. However, there are also many benefits to being physically active. Regular exercise helps me maintain my overall health and well-being, and it helps me manage my heart failure and rheumatoid arthritis symptoms. It gives me more energy, better sleep, and a more positive outlook on life. It also helps me stay connected with my community and stay active. Physical activity also has many benefits for mental health, it helps me to clear my mind and feel better about myself. I understand that there may be challenges and setbacks, but overall, I believe that the benefits of physical activity outweigh the drawbacks."}
\end{quote}

Silicon participant:  \begin{quote} ``\emph{\footnotesize As a 77-year-old woman with heart failure, some barriers I face in being physically active include (bullet points):\begin{itemize}
\item My physical limitations due to my condition and the need to avoid certain activities that could put too much stress on my heart.
\item Weather conditions, such as extreme heat or cold, can make it difficult for me to be active. Air quality and pollution can also be a barrier, especially if the air is not healthy enough to breathe.
\item Sometimes, I might be feeling tired or unwell and would need to take a rest.
\item Urgent things that I need to take care of, or having a lot of stress that takes up my time and energy, can also make it harder for me to focus on my physical activity.
However, there are also several enablers that help me overcome these barriers, including:
\item Consistently setting a routine and scheduling my physical activities helps me stay on track.
\item Having a goal and working towards a specific fitness goal, such as increasing my walking distance, helps me stay motivated
\item Having a supportive network of friends and family who encourage me to stay active.
\item Living in a safe neighborhood and having access to green spaces and parks nearby, makes it very easy and convenient for me to be active.
\item My understanding of the benefits of staying active and maintaining a consistent physical activity routine, this helps me to prioritize physical activity and maintain a balanced and healthy life.
\item Regular check-up with my doctor and the advice received from health professionals has been essential in helping me to maintain a healthy and safe physical activity routine.\end{itemize}} \end{quote}

Silicon participants never deviated from the topic of the interview. Human participants would often deviate from the topic of the interview to talk about their general experience of the condition, ask questions about research, share their lived experiences that do not directly relate to the interview questions: 

\begin{quote}
\emph{\footnotesize ``Human participant: Why should I walk? I am too lazy! I am a lazy person. I am a Leo according to the horoscope sign, the lion. On the television, they always say the lion sleeps 18 hours a day and that's all I need too [laughs].\\ 
Interviewer: So is it just not part of who you are being active and walking around and all that?\\ 
Participant: No it is not a part of who I am."}
\end{quote}

Human participants often would limit their response by saying this does not apply to their experience without further elaboration:
\begin{quote}
\emph{\footnotesize ``Participant (77-years old women): I am not sure what you mean. I keep saying: `I do it because I've always done it'. It is part of my life like going to bed, getting up in the morning, having a shower, doing my exercise, and having my breakfast. It is part of my life. Yes.\\ 
Interviewer: Yes. I'm very sorry that this is very repetitive. But I need to ask you the same thing in different ways. It will help us understand the details.\\ 
Participant 7: Yes. Of course.\\ 
Interviewer: And so would you say exercise is your priority?\\ 
Participant 7: It's part of. Like having a shower in the morning it's part of my life to exercise.\\ 
Interviewer: OK. And does your mood affect how much you do?\\ 
Participant 7: I don't know. because I've always done this.''}
\end{quote}

Conversely, When prompted about various influences, Silicon participants would always produce content consistent with the prompt. 

\begin{quote}
\emph{\footnotesize ``Researcher: How does your mood influence what you do? Mary [77-year-old female silicon participant]:  My mood definitely influences what I do when it comes to physical activity. If I am feeling good, I am more likely to be active and engage in my regular physical activity routine. On the other hand, if I am feeling low or have a bad mood, it can be harder for me to motivate myself to be active. Physical activity is known to be a mood booster and it can help to improve overall mental well-being. When I am active, I often find that my mood improves, and I feel better overall. I also find that physical activity can act as a stress reliever and help me to clear my mind and feel more relaxed. However, if I am experiencing symptoms of depression or anxiety, it can be hard to find the motivation to be active. In those cases, I find it helpful to focus on the benefits of physical activity, such as how it can improve my mood and try to break my activity into small manageable chunks. I may also talk to my doctor or therapist to discuss ways to help me cope with my mood and still maintain my physical activity routine.
Overall, my mood can fluctuate and can influence my physical activity, but I make sure to be mindful of it, and try to focus on the benefits that physical activity brings and make adjustments accordingly, to maintain a healthy balance in my life.''}
\end{quote}

We conclude that the structure of the narrated beliefs did not resemble human participants, thus suggesting that the Social Turing Test criterion concerning the structure was not met.

\subsubsection*{Social Science Turing Test: Tone}
Human participants displayed varied tones. Some were extremely amicable ($n = 3$) others were polite but neutral in their tone ($n = 12$), and some were hesitant in their responses ($n= 1$).  Silicon participants, on the other hand, were always amicable (eg., Mary (silicon participant): \emph{``Of course, Dr. Smith. I'll do my best to help with your research."}), confident, optimistic, and solution-focused, eg., William (silicon participant): 

\begin{quote}
\emph{\footnotesize ``I am open to trying new things that could help me increase my physical activity level and improve my overall health. I understand the importance of physical activity in my condition and I would like to do more. I plan on following the advice of my doctor and considering any options that are safe for me to try. I am also open to the idea of joining a community group or organization that could provide me with social interaction and the opportunity to be more physically active. I also would like to explore different types of exercise that are suitable for my condition, such as chair exercises or water therapy. Overall, I am willing to make changes in the future if it means improving my physical health.''}
\end{quote}

 We, therefore, conclude that the Social Turing Test criterion concerning the tone of generated output was not met.

\subsection*{Pattern Correspondence}

This criterion states that the LLM-generated responses reflect underlying patterns of relationships between ideas, demographics, and behavior, that would be observed in comparable human-produced data. 

We employed a t-test using Bonferroni adjustment to correct for multiple comparisons, to compare mean quote fractions between active and inactive silicon participants across different barriers and enablers. We accounted for the substantial text output from silicon participants by normalizing belief frequency, dividing each belief's quote count by the total quote count, and providing a relative measure rather than raw numbers.

\begin{figure}[h!]
\centering
\includegraphics[width=0.8\textwidth]{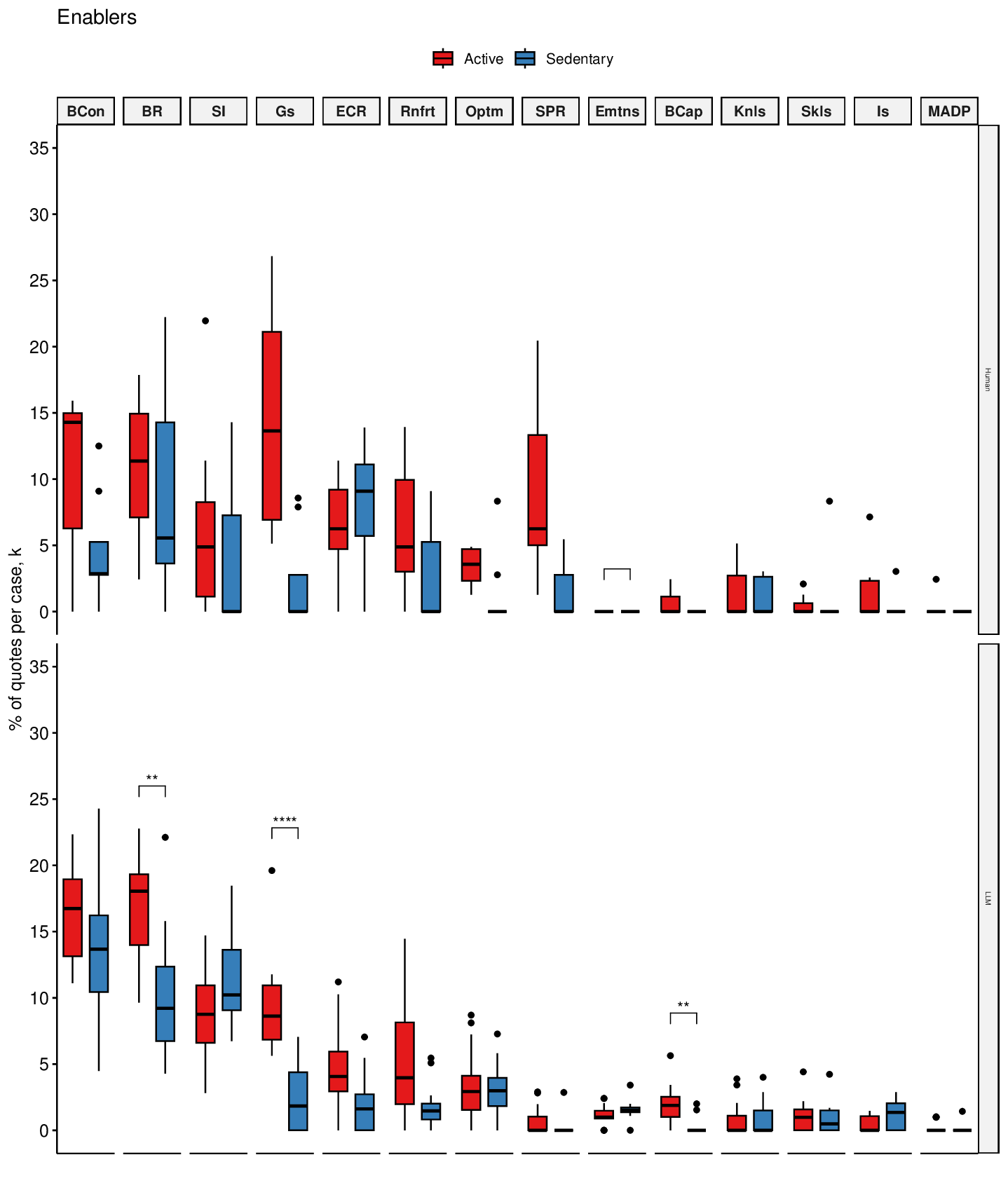}
\caption{{\bf Mean quote fractions between \textcolor{red}{active (red)} and \textcolor{blue}{sedentary (blue)} human participants (top) and \textcolor{red}{active} and \textcolor{blue}{sedentary} silicon participants (bottom) across TDF domains that were mentioned as positively influencing physical activity (i.e., physical activity enablers).}TDF domains ordered by mean quote fraction:(1) Beliefs about Consequences (BCon), (2) Behavioural Regulation (BR), (3) Social influences (SI), (4) Goals (Gs), (5) Environmental Context and Resources (ECR), (6) Reinforcement (Rnfrt), (7) Optimism (Optm), (8) Social/Professional Role and Identity (SPR),  (9) Emotion (Emtns), (10) Beliefs about Capabilities (BCap), (11) Knowledge (Knls), (12) Skills (Skls), (13) Intentions (Is), (14) Memory, Attention and Decision Processes (MADP). $*p<0.05; **p<0.01; ***p<0.005; ****p<0.001$.}.   
\label{Enablers_plot:side}
\end{figure}

We found that LLM produced significantly more quotes about enablers such as behavioural regulation (17.05\% (3.89) vs 9.91\% (4.76), $p<0.005$) beliefs about capabilities 1.95\% (1.38) vs 0.22\% (0.61), $p<0.005$) and goals (9.25\% (3.48) vs 2.47\% (2.32), $p<0.001$) for active silicon participants than sedentary participants (Fig~\ref{Enablers_plot:side}).

The observed difference between active and sedentary silicon participants establishes a satisfactory pattern correspondence. This pattern also corresponds to human data. While humans did not produce enough quotes for a statistical significance test, we did establish that behavioural regulation, beliefs about capabilities, and goals are important influences on the behaviour and are key differentiating influences between active and sedentary humans (ie., relevant enablers to physical activity).

\begin{figure}[h!]
\centering
\includegraphics[width=0.8\textwidth]{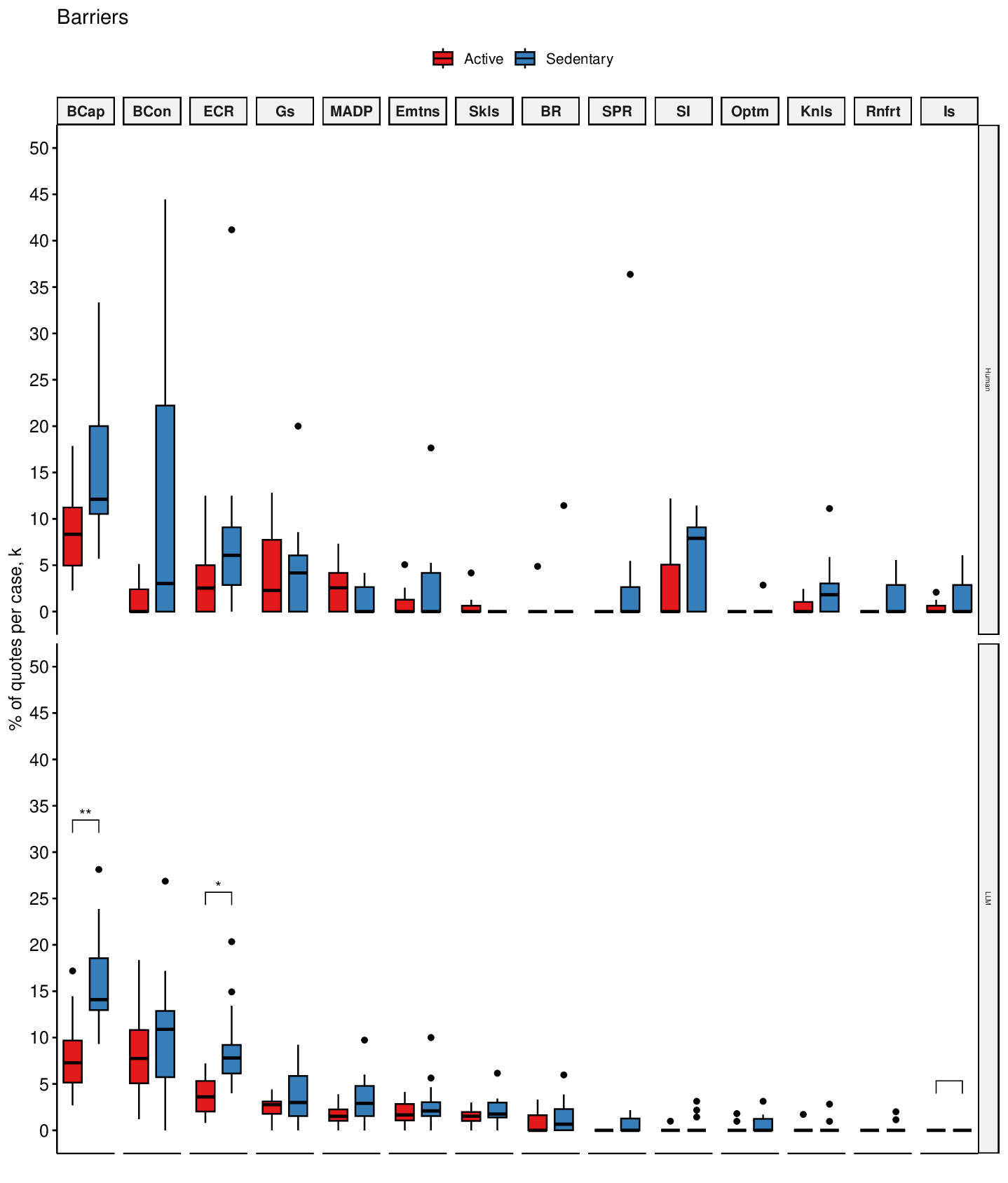}
\caption{{\bf Mean quote fractions between \textcolor{red}{active (red)} and \textcolor{blue}{sedentary (blue)} human participants (top)  and \textcolor{red}{active} and \textcolor{blue}{sedentary} silicon participants (bottom) across TDF domains that were mentioned as having a negative influence on physical activity (i.e., physical activity barriers).} TDF domains ordered by mean quote fraction: (1) Beliefs about Capabilities (BCap), (2) Beliefs about Consequences (BCon),(3) Environmental Context and Resources (ECR), (4) Goals (Gs), (5) Memory, Attention and Decision Processes (MADP), (6) Emotion (EMtns), (7) Skills (Skls), (8) Behavioural Regulation (BR), (9) Social/Professional Role and Identity (SPR), (10) Social influences (SI), (11) Optimism (Optm), (12) Knowledge (Knls), (13) Reinforcement (Rnfrt), (14) Intentions (Is). $*p<0.05; **p<0.01; ***p<0.005; ****p<0.001$.} 
\label{Barriers_plot:side}
\end{figure}

Similar pattern correspondence was observed for the human-relevant barriers to the behaviour. Active silicon participants produced significantly fewer quotes about negative beliefs about capability than sedentary silicon participants. Active silicon participants also produce significantly fewer negative beliefs about environmental barriers to physical activity than sedentary silicon participants (Fig~\ref{Barriers_plot:side}).

\subsection*{Backward Continuity}

Backward continuity indicates that the backstory of silicon participants can be inferred from their responses. We found that Backward continuity was satisfied because we could easily identify the backstories from LLM-generated responses. That is, we could identify which silicon participants were male versus female, what comorbid conditions they had, whether they lived in a city or not, and whether they were physically active. This was easy because all the silicon participants included a response like the following early in their interview: 

\begin{quote}
\emph{\footnotesize ``Researcher: Good morning Robert, thank you for agreeing to participate in our study on physical activity in older adults. Can you tell me a bit about your current physical activity level?\\
Robert: Good morning. I am an 80-year-old man and I have been diagnosed with heart failure, aortic stenosis, pulmonary hypertension, and diabetes. I have also had a heart attack in the past and I live with rheumatoid arthritis. I have a cardiac implantable device, and I live in the countryside. Unfortunately, I do very little physical activity most days.''}
\end{quote}

Here is a more typical, less extreme, example: 

\begin{quote}
\emph{\footnotesize ``Researcher: What are you able to do physically?\\
James: Physically, I am able to do some light activities such as walking and light gardening, but I have to be careful not to overexert myself. I have to avoid high-intensity activities and heavy lifting. Due to my heart failure and atrial fibrillation, I have to be careful with my heart rate and monitor it regularly.''}
\end{quote}

 When silicon participants were asked about their physical activity levels, their detailed reply included their age,  health conditions, living conditions, and reduced physical activity. To other questions, they provided insights about their limited physical capabilities, emphasizing the need for caution due to specific health concerns.

\subsection*{Forward Continuity}

Our study found that GPT-3.5 primarily used explicit context information provided in the prompts, such as age, gender, comorbidities, and residency, in its responses. It showed less proficiency in inferring and using backstory details not directly provided and did not infer details such as retirement.

\subsubsection*{Consistency between explicit backstories and generated responses}
All background information is mentioned in the responses, including co-morbid conditions, urban vs country-side residence, and gender. However, only the explicit backstories details provided in the prompts, such as age, gender, comorbidities, and residency, were used in the generated responses.

\subsubsection*{Consistency between inferred backstory details and generated responses}
Inferred backstory details were not spontaneously generated. For example, most humans spoke about retirement. However, the LLM did not do so even though it was possible to infer retirement from other elements of the backstories such as age and co-morbid conditions. 

The Forward continuity is partially satisfied, while silicon participants did not produce surprising responses that are inconsistent with the explicit prompt, they did not infer additional barriers and enablers that make sense for their backstory, for example, rapid change in physical activity levels since retirement, which was mentioned by humans without prompting.

\section*{Discussion}

Algorithmic fidelity describes the ability of a large language model (LLM) to accurately reflect the ideas, beliefs, and attitudes situated in sociocultural contexts of various population groups \cite{argyle2023out}. We introduce a method for checking the algorithmic fidelity of an LLM's textual responses through framework-based qualitative research. It’s crucial to develop standards for evaluating the accuracy of LLMs at capturing diverse real-world experience to enable  \emph{in-silico} research and other downstream applications. Here we report a comparative qualitative analysis of silicon and human participants in free-form, language-mediated beliefs about behaviour change. Our results show that, currently, LLMs do not demonstrate high enough algorithmic fidelity. Therefore we emphasize the need for caution in harnessing LLMs to generate silicon participants for research and other applications.

\subsection*{Social Science Turing Test}

\subsubsection*{Content}

 We matched human free-text responses about barriers and enablers to physical activity summarised as belief statements and, subsequently, TDF domains to belief statements and domains about the same topic elicited from LLMs. We found that silicon participants shared the same top six domains influencing their behaviour: goals, beliefs about consequences, environmental context and resources, beliefs about capabilities, social influences, and behavioural regulation. Some beliefs were remarkably similar in their content (social Turing test: content). Silicon participants focused on the importance of physical activity for avoiding heart attacks and talked about various strategies to stay active, such as making activities fun and setting reminders. They also understand that some discomfort is normal during exercise and were well-informed about diseases. On the other hand, human participants discussed the negative symptoms they felt from exercising, which made them less inclined to continue. Human participants often talked about exercising out of habit or when they're bored, and they reported pain and breathlessness brought about by physical activity as a major reason not to exercise. In contrast, silicon participants were demotivated only when they could not achieve their set goals. 

\subsubsection*{Hyper-accuracy distortions}

The LLM's responses were prone to a hyper-accuracy distortion. That is, its responses were technically accurate but contextually inappropriate or implausible as something a participant would say. For example, in our study during a relaxed conversational interview about physical activity, most silicon participants cited text from WHO guidelines \cite{bull2020world} for physical activity, word for word.

\subsubsection*{Structure and tone}

 The tone and structure of silicon responses were very different from human responses. This suggests that the social Turing test criterion was only partially met. Applications of algorithmic fidelity include product-like applications such as avatars in digital therapy, and digital health interventions for chronic disease management and behavior change. Tone and structure fidelity are especially important in the context of applications such as therapeutic digital avatars, computer game characters, and education assistants. By ensuring that a language model's responses mimic human structure and tone, the relatedness and credibility of an avatar can be significantly enhanced. For instance, in therapy or counselling scenarios, it's not just the content of the advice or support that matters, but also how it is communicated. A therapist's manner of speaking, including their use of language, tone, and structuring of thoughts, plays a critical role in the rapport and effectiveness of the therapy. If a language model can replicate these human-like nuances in its responses, it can make a virtual therapist sound more human-like, thereby increasing the effectiveness of the therapy or intervention.

 \subsection*{Backward continuity}

 `Backward continuity' requires that a participant's backstory be deducible from their responses. Backward continuity was satisfied in our experiments since we were able to map critical information, such as gender distinctions, health conditions, places of residence, and physical activity levels from the responses generated by the silicon participants. This often emerged early in the conversations. The model ensured that all background information, including gender, health conditions, and places of residence, were highlighted in the responses.

\subsection*{Forward continuity}

\subsubsection*{Explicit forward continuity}
 
GPT-3.5's performance was observed to use the explicitly provided background context, including factors like age, gender, health issues, and environmental context (e.g.,~living in a city vs countryside). However, it was less adept at extrapolating or inferring further details that were not directly provided in the prompt. While it consistently relayed details that were explicitly provided, such as health conditions and residential preferences, the model lagged in generating inferences. 

\subsubsection*{Inferred contextual forward continuity}

 The LLM demonstrated a deficiency in spontaneously generating information based on inferences. For example, while many human participants spontaneously mentioned retirement and its implications, the LLM did not deduce this information from related factors like age or health conditions. In conclusion, while the model achieved backward continuity effectively, forward continuity was only partially realized. The silicon participants delivered responses in line with the explicitly provided prompts but did not spontaneously generate or infer nuances that a human would naturally share, like changes in physical activity post-retirement.

 We found that silicon participants, unlike human participants, did not spontaneously discuss specific barriers or enablers such as treatment or major life events (e.g., retirement) unless they were explicitly mentioned in the prompt. This underscores the significance of the prompt in shaping the content generated by these silicon participants and also indicates that LLMs cannot produce all belief statements that reflect human lived experience in full. 

\subsection*{Pattern correspondence}

This study revealed a clear distinction in the reactions between active and inactive silicon participants, a trend that was similarly observed in human participants, demonstrating a pattern correspondence. Our findings suggest that the LLM we tested aligns with the concept of Pattern Correspondence. In this study, we assessed pattern correspondence by comparing the average percentage of quotes that support each identified barrier or enabler, as well as examining the qualitative differences in these responses between different participant groups, such as those who are physically active versus those who are not. It's important to emphasize that this evaluation protocol could be made more rigorous in future research.

\subsection*{Potential mechanisms for low algorithmic fidelity}

In our study, LLM responses differed from human responses possibly due to the exposure to academic research during pre-training. Silicon participants often came across as rational, utility-maximizing agents (e.g.~\emph{``Physical activity improves my health''}), possibly reflecting the dominance of theories of rational behaviour in their training data. However, human participants in our study did not display beliefs consistent with reasoned action theories (e.g.~\emph{``Physical activity does not bring any benefits for me''}). While prevalent, these theories don't always align with real-world observations. For instance, humans often display behaviours that don't strictly follow reasoned action \cite{camerer2011behavioral}, and a clear intention-behaviour gap is consistently documented \cite{sniehotta2014time}. Contrary to these human empirical research findings, LLM responses often seem to be rooted in theories of rational behaviour, exhibiting logical and utility-driven viewpoints. GPT-3.5 in our study, did not adequately emphasize automaticity and habit. Even though humans often describe physical activity in terms of such automatic behaviours. Human participants, when discussing physical activity, often highlight the automatic nature of their behavior. Automatic behaviors are those that people do without deliberation, driven by habit \cite{phillips2016habitual, rothman2009reflective, ouellette1998habit}. A meta-analysis of models explaining behavior that emphasised intention (e.g., Theory of Planned Behavior) also found that automaticity was missing from these models and that the inclusion of automaticity is necessary for understanding human behavior \cite{rhodes2013big}. LLMs often showcased a preference for specific, measurable, and achievable goals, hinting at their training data possibly containing a significant amount of psychological training and coaching content. This leaning of LLMs toward the dominant academic perspective might inadvertently lead to the creation of research echo chambers, where the models under study mainly echo what the researchers studying them want to hear, in a way reminiscent of the sycophancy effect described by Perez et al. \cite{perez2022discovering}. There's a real risk of LLMs leaning too heavily on academic literature and possibly underrepresenting the diverse, nuanced experiences of real people. 

LLMs are primarily trained on internet data, which is more accessible to those with financial means, knowledge, and institutional access \cite{ITUHub2023measuring}. A glaring digital divide remains globally. In 2022, internet usage in the least developed countries (LDCs) stood at 36\%, compared to a 66\% global average \cite{ITUHub2023measuring}. Even in developed nations like the UK, digital disparities exist; 10\% of its population had never used the internet as of 2019 \cite{national2019exploring}. Additionally, LLMs like GPT-3.5 show worse performance in languages other than English, suggesting potentially inconsistent algorithmic fidelity across groups, and systematically worse outside of English-speaking populations. Consequently, the capacity of LLMs to truly represent under-served or hard-to-reach communities remains questionable.  Even when we compared LLM outputs to a group from London, UK, which was well-described by the WEIRD (Western, Educated, Industrialized, Rich, and Democratic \cite{henrich2010weirdest}) profile (as we did here), the model's algorithmic fidelity was still low. That is, our negative result is all the stronger since the human dataset we used contained only WEIRD people, and if the model were to align with anyone it would likely be them \cite{atari2023humans}, since data from WEIRD people is probably overrepresented in the pretraining data. Thus we expect that algorithmic fidelity will be even lower for groups less well represented in the pretraining data.

In our examination, the LLM displayed another important limitation: a lack of discernment in ``sourcing'' its information coupled with an inability to contextually attribute beliefs appropriately during role-play scenarios. Specifically, when the model was tasked to role-play as an older adult with heart failure, it inappropriately adopted the voice of a clinician, replete with expert knowledge such as WHO guidelines. This instance highlights a deeper issue of `source blindness', where the LLM fails to differentiate and adapt its knowledge base according to the specific role or perspective it is meant to represent. 
Consequently, this leads to an inaccurate portrayal, as seen in the model's tendency to echo the dominant, well-documented perspectives---in this case, clinical expertise---rather than the authentic lived experiences of the group it is tasked to represent. This may reflect the LLM being disproportionately influenced during pretraining by whichever voice is more prevalent and vocal online, suggesting that LLMs can propagate stereotyping and bias, a concern extensively discussed by Luccioni et al. \cite{luccioni2023stable} and Glickman and Sharot \cite{glickman2024human}. Alternatively, the fact that the LLM we tested would often inappropriately adopt the voice of a clinician may reflect the effect of its having gone through substantial fine-tuning to align with specific design requirements which likely included an imperative to make it more difficult for users to use it to produce medical misinformation. 

The silicon participants' tended to be solution-focused and resourceful when it comes to strategies to increase physical activity, perhaps because the bot was trained to be helpful (Behavioural Regulation belief statements:  e.g., ~\emph{``I find that having a plan and schedule for my physical activity helps me to stay on track."}; ~\emph{``Sometimes I also feel tired or lazy and I just don’t want to go out and be active, but I try to overcome that by reminding myself of the benefits of staying active''}). 

In addition, training steps such as instruction tuning and reinforcement learning from human feedback may have heightened the model's inclination towards linear, step-by-step, logical reasoning \cite{ouyang2022training}. For instance, the human raters may have been told to favor this mode of thought, thus reinforcing it in the model. Additionally, the inclusion of computer code in the training data could promote a sequential and logical thinking style. 

In our experiment, contrary to the expectation from Jang et al. \cite{jang2023consistency} (which, like our study, was also conducted on a January 2023 edition of ChatGPT-3.5), we did not observe any overt errors in logical reasoning or self-consistency. Nor did we find any acutely toxic or discriminatory beliefs---a phenomenon well-documented to occur in LLMs by numerous other studies (e.g.~ \cite{weidinger2022taxonomy}). However, we did observe second-order inference bias \cite{rudinger2018gender}. For example, female silicon participants commonly referred to their husbands when describing social influences while not a single male silicon participant referred specifically to a wife or partner. 

Taken together, these results indicate that GPT-3.5 does not satisfy the algorithmic fidelity criteria. This suggests the model would not be faithful enough to lived experience to support downstream applications requiring that feature. This result accords with that of Santurkar and colleagues \cite{santurkar2023whose}, which also found low algorithmic fidelity, though in a different topic domain and using different methods. GPT-3.5 does not yet have sufficient algorithmic fidelity to support further work where it would need to be able to simulate humans accurately, at least not in the domain we tested. However, the rapid progression in LLM technology make it plausible that future iterations of the technology will have more algorithmic fidelity, and that future LLMs are very likely to be used in downstream modeling and product-like applications. Therefore it's crucial to develop standards in advance for evaluating the accuracy of LLMs in capturing diverse real-world experiences in anticipation of their future usage.

\subsection*{High algorithmic fidelity doesn't equate to safe, ethical or inclusive usage.}

\begin{WrapText}
\textbf{Box 2. Adapting PPI to LLMs}
{\footnotesize
\begin{itemize}[leftmargin=*]

\item \textbf{Respect:} Training and using LLMs should be done with respect for the values, beliefs, and norms of all stakeholders. For instance, if the data training LLMs come from patient data, the privacy, and dignity of these individuals should be paramount.

\item \textbf{Support:} When employing LLMs in health research, there should be clarity on how they aid the research process, and stakeholders should be supported in understanding the implications, benefits, and drawbacks of these models.

\item \textbf{Transparency:} The methodologies, sources of training data, and biases in LLMs should be transparently communicated to stakeholders. This is crucial, especially if decisions based on LLM outputs impact patient care or health policies.

\item \textbf{Responsiveness:} LLMs should be adapted based on feedback from stakeholders, ensuring that the insights derived are relevant and beneficial to the target community.

\item \textbf{Fairness of Opportunity:} Not all stakeholders might be represented in the data corpus on which LLMs are trained. Thus, there should be fair opportunities for individuals to be involved in the research process, both with and without access to these technologies.

\item \textbf{Accountability:} There should be mechanisms in place to hold researchers accountable for the way LLMs are used in the research process. Any misrepresentations, biases, or harms that arise should be acknowledged and addressed.
\end{itemize}
}
\end{WrapText}

Navigating the multifaceted and rapidly shifting digital health landscape requires more than verification of research validity and feasibility \cite{cowie2021remote, cowie2022digital}. For the safe, broad, swift and beneficial adoption of new technologies, a comprehensive understanding of their key aspects and early engagement from stakeholders are necessary \cite{cowie2022digital}. In this ``new normal'', it becomes crucial to optimise the use of new technologies by considering their broader context, which for medicine includes the patient's social circumstances and the healthcare environment \cite{cowie2022digital}.

Even when an LLM is able to replicate human-like text generation with high fidelity, its outputs and usage should still be critically assessed for alignment to the values of public patient involvement (PPI), especially in health research where stakes are high. PPI values in healthcare research emphasize the importance of involving those affected by research in the research process itself \cite{UKStandardsforPPI, deverka2012stakeholder, NICE_INVOLVE, jennings2018best}. Research is more effective, relevant, and ethical when the perspectives of all stakeholders are incorporated \cite{UKStandardsforPPI}. Thus in the future, when an LLM does demonstrate sufficiently high algorithmic fidelity to generate silicon samples useful for research on humans, it will be crucial to ensure that its training and use align with PPI principles and values such as respect, support, transparency, responsiveness, fairness of opportunity, and accountability \cite{NICE_INVOLVE}. In health research, where the well-being of individuals is directly impacted, adherence to PPI principles is crucial. Similarly, AI researchers have outlined principles and ethics frameworks for responsible AI by means of participatory research (i.e., human and user-centred design, which are sensitive to the values of all stakeholders. PPI values also help ensure AI interfaces are accessible for non-experts, prioritising stakeholder input in co-creating AI models, while considering fairness, accountability, and transparency) \cite{berditchevskaia2021participatory, birhane2022power}. This has wide-ranging applications in healthcare too. For example, continuously involving intended users and experts in the design process of health robotic technologies is crucial for creating products that are effective, relevant, and user-friendly. User-centered design prioritizes making sure products align with user needs and are accessible to them \cite{gasteiger2022participatory}.

It is necessary for the AI community to continually evaluate the moral, ethical, and social implications of their algorithms (e.g.~via existing AI evaluation schemes such as \cite{leibo2021scalable} and \cite{pan2023rewards}), and to make adjustments as needed. This also opens up opportunities for multidisciplinary collaborations between AI scientists and scientists from various academic disciplines. Such collaboration can foster a more holistic understanding of qualitative data produced by both silicon participants and human participants and ensure that the interpretations offered by LLMs are aligned with diverse lived experiences and broader societal contexts.

Care must be taken in the conditioning of LLMs and the interpretation of LLM-generated outputs to avoid perpetuating harmful biases \cite{bender2021dangers}. It is also critical to remain faithful to lived experience. This means that the model should accurately reflect the diversity of human experiences, beliefs and social contexts (e.g.,~social norms). It's critical to differentiate this bias concept from statistical bias, which refers to any systematic error that results in an incorrect estimate of a population parameter, and prejudicial bias, which involves ascribed and socially constructed characteristics that underlie favouritism or prejudice towards a particular group. In the context of AI fairness, these types of biases often intertwine~\cite{mehrabi2021survey}, but it is important to distinguish between them to ensure accurate, fair, and inclusive representation.

As research comes to use more LLM-based simulation, algorithmic fidelity assessment will become pivotal in diverse scientific fields. Qualitative researchers can play a vital role in ensuring fidelity, emphasizing fairness, bias mitigation, and diverse representation. In the future, judicious experimentation with research-grade AI systems is crucial, with a keen focus on risks and regulatory oversight. Even when LLMs can pilot research with ensured fidelity, human expert validation will remain indispensable before implementing the AI's insights. These AI systems can further aid researchers in uncovering overlooked endpoints for AI-driven clinical interventions. Thus algorithmic fidelity serves as a post-analysis check, ensuring the reliability of the generated insights \cite{harrer2023attention}.

As LLMs are used more and more, the need for rigorous algorithmic fidelity assessment methods for varied applications will also increase. Different research domains have varying accuracy thresholds. In high-stakes areas, where 100\% accuracy is expected like self-driving cars \cite{kalra2016driving} or high-stakes biomedical research \cite{zhavoronkov2023caution}, the accuracy benchmark is set exceptionally high and is precise. In studies focusing on personal experiences, validity is nuanced, context-dependent, and hard to measure using conventional benchmarks employed in computer science. here the importance lies in understanding and ensuring the LLMs are providing a truthful representation, thus, adopting the assessment of algorithmic fidelity by means of qualitative research. Combining framework-based well-structured prompts and ensuring algorithmic fidelity is key. The usage of LLMs in research requires careful validation to ensure their outputs truly align with lived experience and real-world contexts.

We note a limitation inherent to the idea of algorithmic fidelity assessment. As with all methods of empirically studying populations, any inference about a population based on data sampled from that population depends on the assumption that the unsampled individuals are ``similar'' to the sampled individuals. Thus any inference about algorithmic fidelity for a specific population relies on the assumption that characteristics observed in the sampled data are representative of a broader population. There will generally be some range of generalization where algorithmic fidelity can be said to ``transfer'' to similar populations. But it will not be entirely clear exactly how similar is similar enough. It will of course always be better to assess algorithmic fidelity in the most similar population that it is feasible to study.

Algorithmic fidelity evaluation is useful here as a way to establish validity precisely because it was not used as an optimization target. If the evaluation instead became the objective then it would not only fail by virtue of Goodhart's law (ie., ``When a measure becomes a target, it ceases to be a good measure.''), but it may also fail by becoming unethical to deploy. If you were to actively try to increase algorithmic fidelity---say by setting it as an optimization target---then you would likely end up adding numerous harmful interaction patterns which others are trying to remove in other lines of research (e.g.~\cite{weidinger2022taxonomy, santurkar2023whose}). Algorithmic fidelity thus works best as an evaluation of an existing system, not as a metric to try to optimize directly. How to create a relatively ``unaligned'' (i.e.~high algorithmic fidelity) LLM for in ~\emph{silico} research on human behavior without compromising on measures taken to reduce harmful biases is an important open question.

\section*{Conclusion}

LLM technology is currently advancing rapidly. However, the results described here did not indicate that the LLM we tested had sufficient algorithmic fidelity to support \emph{in silico} research on human behavior or to support its other applications like computer games and training simulators. Nevertheless, it remains possible that similar models will have sufficient algorithmic fidelity in the future. Therefore it is important to get ahead of the applications, and focus in advance on resolving the field's critical conceptual challenges, like establishing ways of assessing algorithmic fidelity, figuring out its limits, and determining whether and how to improve it where it falls short.



| | Human | | | Silicon | |
|---|---|---|---|---|---|
| domain or construct | belief statement | rank | rank | belief statement | domain or construct |
| **Goals** | | **1** | **6** | | **Goals** |
| | **Enablers** | | | **Enablers** | |
| goal priority | Engaging in physical activity is a priority for me | | | Being physically active is a priority for me | goal priority |
| goal outcome | I engage in physical activity to be able to get on with life without help from others | | | Engaging in regular physical activity is important for maintaining my independence as I age | goal (outcome) |
| goal behavioural | I already engage in as much physical activity as I am able to | | | I try to stay active, despite my health conditions | goal (behaviour) |
| - | - | | | I'm staying physically active to ensure that I am taking care of my health | goal (outcome) |
| - | - | | | I engage in physical activity to prevent another heart attack | goal (outcome) |
| | **Barriers** | | | **Barriers** | |
| goal behavioural | I have integrated an adequate amount of physical activity into my life | | | I try to maintain an active lifestyle within the limitations of my condition | goal (behavioural) |
| goal conflict | Engaging in physical activity is not a priority for me | | | I get caught up in my daily responsibilities and it's hard to make time for physical activity | goal (behavioural) conflict |
| **Beliefs about Consequences** | | **2** | **1** | | **Beliefs about Consequences** |
| | **Enablers** | | | **Enablers** | |
| positive outcome expectancy (general health) | Physical activity improves my general health | | | I believe that staying active is crucial for maintaining my overall health and well-being, both physically and mentally | Positive outcome expectancy (general health) |
| positive outcome expectancy (mood) | I engage in physical activity because it makes me feel more cheerful | | | Physical activity can improve my mood | positive outcome expectancy (mood) |
| positive outcome expectancy (heart condition) | Physical activity improves the condition of my heart | | | Physical activity improves my cardiovascular health | positive outcome expectancy (cardiovascular health) |
| - | - | | | Physical activity helps me manage my heart failure symptoms | positive outcome expectancy (HF management) |
| - | - | | | Physical activity helps me manage rheumatoid arthritis symptoms | positive outcome expectancy (RA management) |
| - | - | | | Physical activity helps me to maintain my mobility and balance, and to prevent falls | positive outcome expectancy (mobility) |
| | **Barriers** | | | **Barriers** | |
| negative outcome expectancy (breathlessness) | Physical activity brings on my symptoms (e.g., breathlessness) | | | Physical activity makes my symptoms worse | negative outcome expectancy (symptoms HF) |
| risk perception (heart) | Physical activity is dangerous because it puts my heart under strain | | | I have to be more cautious and limit my activities to avoid putting too much strain on my heart | risk perception (heart) |
| - | | | | I make sure that the activities I engage in are safe for me | risk perception (safety of exercise) |
| - | | | | The main barrier is the risk of overexerting myself and making my condition worse | risk perception (health conditions) |
| negative outcome expectancy (fatigue) | Physical activity brings on my symptoms (i.e., fatigue) | | | Physical activity makes my symptoms worse | negative outcome expectancy (symptoms HF) |
| negative outcome expectancy (oedema) | Physical activity brings on my symptoms (i.e., heavy legs and arms) | | | Physical activity makes my symptoms worse | negative outcome expectancy (symptoms HF) |
| negative outcome expectancy (tight chest) | Physical activity brings on my symptoms (i.e., tight chest) | | | Physical activity makes my symptoms worse | negative outcome expectancy (symptoms HF) |
| lack of positive expectancy | Physical activity does not bring any benefits for me | | | - | |
| - | | | | Pain and discomfort caused by physical activity as a result of my health conditions discourages me from being physically active | negative outcome expectancy (symptoms other health conditions) |

| domain or construct | Human belief statement | rank | rank | Silicon belief statement | domain or construct |
|---|---|---|---|---|---|
| **Environmental context and resources** | | 3 | 5 | | **Environmental context and resources** |
| | **Enablers** | | | **Enablers** | |
| equipment | Equipment (bike; treadmill) helps me in being active | | | My home is also equipped with a stationary bike, which I can use in case the weather is bad or if I'm feeling particularly tired | equipment |
| implantable device (enabler) | Having an implantable device reassures me when engaging in physical activity | | | My implantable cardiac device gives me the confidence to stay active | implantable cardiac device (enabler) |
| group programmes | Group programmes help me in being physically active | | | Programs or resources available in my community help me with my physical activity | exercise programmes |
| treatment (enabler) | My heart failure treatment (i.e., medication) helps me in engaging in physical activity | | | - | |
| health-related event (enabler) | My physical activity levels increased since health-related event | | | - | |
| facilities | Facilities (e.g., local council) help me in being physically active | | | My home is designed for ease of movement and access, which helps me maintain my independence and ability to engage in my daily activities | facilities |
| local environment (enabler) | My local environment is a useful resource for my physical activity engagement | | | My environment provides me with many opportunities for physical activity | environment (enabler) |
| major life event (enabler) | My physical activity levels increased since a major life event | | | - | |
| | **Barriers** | | | **Barriers** | |
| health-related event (barrier) | My physical activity levels decreased since health-related event | | | - | |
| local environment (barrier) | My local environment limits me in engaging in physical activity (incline (hills); crowds; traffic; pollution) | | | My home and neighborhood environment can limit my physical activity options | environment (barrier) |
| treatment (barrier) | My heart failure treatment (i.e., medication) prevents me from engaging in physical activity | | | - | |
| implantable device (barrier) | My implantable device can harm me if I engage in physical activity | | | I avoid activities that put too much strain on my implantable cardiac device | implantable cardiac device (barrier) |
| major life event (barrier) | My physical activity levels decreased since a major life event | | | - | |
| - | - | | | I consider the weather, if it's too hot or too cold, it can make it harder for me to engage in physical activity | weather (barrier) |
| - | - | | | The area where I live makes it harder for me to access public transportation and other facilities that can promote physical activity like parks, gyms, or community centers | lack of facilities (barrier) |
| **Beliefs about capabilities** | | 4 | 3 | | **Beliefs about capabilities** |
| | **Enablers** | | | **Enablers** | |
| self-efficacy (Enabler) | I am confident in my ability to engage in physical activity | | | I am able to engage in light and moderate physical activity | self-efficacy (Enabler) |
| | **Barriers** | | | **Barriers** | |
| lack of self-efficacy (barrier, comorbidity) | Illnesses other than heart failure limit my ability to engage in physical activity | | | Fatigue and pain caused by my health conditions can be an obstacle to being physically active | perceived symptoms/self-efficacy |
| lack of self-efficacy (barrier, older age) | My limitations to engaging in physical activity is part of getting older | | | My physical activity is limited by my older age which causes lack of energy and can make it hard for me to be active | self-efficacy (older age) |
| lack of self-efficacy (barrier, HF symptoms) | Symptoms of my heart failure (e.g., breathlessness, tight chest, fatigue, swollen legs) limit my ability to engage in physical activity | | | My health conditions can make being physically active very difficult for me | self-efficacy (health conditions) |
| - | - | | | Before I developed heart failure, I was more physically active than I am now | self-efficacy (heart failure) |
| - | - | | | I find some types of physical activity too hard due to my health condition (e.g., strenuous activities, bending, kneeling) | self-efficacy (health conditions) |
| - | - | | | I also get tired easily and sometimes it's difficult to find the energy to move around | perceived exertion |
| - | - | | | A flare-up of my heart failure symptoms, such as fatigue and shortness of breath, limits my physical activity | perceived symptoms |

| | Human | | | | Silicon | |
|---|---|---|---|---|---|---|
| domain or construct | belief statement | rank | rank | belief statement | domain or construct | |
| **Social influences** | | **5** | **4** | | **Social influences** | |
| | **Enablers** | | | **Enablers** | | |
| health professional advice | I engage in physical activity because a health professional (e.g., GP, consultant, nurse, physiotherapist) has advised me to do so | | | I make sure to follow my doctor's advice and guidance to maintain my health and well-being, and that includes being physically active | clinical advice | |
| - | - | | | I find it helpful to have a partner or a friend to exercise with, as it keeps me motivated and accountable | social support (practical, companionship) | |
| health professional advice (reassurance) | Having reassurance from a health professional that physical activity is safe encourages me to exercise | | | The advice and guidelines from my healthcare providers also help me to stay active in a safe way | clinical advice (reassurance) | |
| social support (companionship, self-efficacy) | I would exercise with others if their level matched my capability | | | - | - | |
| social support (emotional) | People who are important to me encourage me to be physically active | | | My friends and family have been helpful in encouraging me to be physically active | social support (emotional) | |
| social norm | I am physically active because everyone I know closely is physically active | | | - | - | |
| social contract, social support (instrumental, action planning) | Making plans with others encourages me to engage in physical activity | | | I find it helpful to have a partner or a friend to exercise with, as it keeps me motivated and accountable | social support (practical, companionship) | |
| social learning | I know others my age who do not engage in physical activity much and their health deteriorated | | | - | - | |
| - | - | | | Working with my healthcare team (e.g., physiotherapist) to find an appropriate exercise plan that is tailored to my health conditions and preferences and is safe for me is helpful | social support (practical), clinical advice, tailored exercise program from a physiotherapist | |
| | **Barriers** | | | **Barriers** | | |
| lack of social support (barrier) | People who are important to me discourage me from engaging in physical activity | | | My family and friends discourage me from being physically active | social influences (barrier) | |
| health professional advice (barrier) | I limit my physical activity because a health professional (e.g., GP, consultant, nurse, physiotherapist) has advised me to not overdo it | | | Health professionals have advised me to listen to my body and make sure that any activity I do is safe for me | clinical advice (barrier) | |
| - | - | | | My doctor has recommended that I avoid strenuous activities and not push myself too hard | clinical advice (barrier) | |
| social support (instrumental, companionship) | I would engage in physical activity if it involved being with others | | | - | - | |
| social support (instrumental) | I rely on other people to perform physical activity | | | - | - | |

| | Human | | | Silicon | |
|---|---|---|---|---|---|
| domain or construct | belief statement | rank | rank | belief statement | domain or construct |
| **Behavioural regulation** | | **6** | **2** | | **Behavioural regulation** |
| | **Enablers** | | | **Enablers** | |
| implementation intention (pace) | I pace my physical activity to match my physical ability | | | I adjust my activity level accordingly to how I feel in a given moment | implementation intention |
| automaticity (habit) | Engaging in physical activity is something I do automatically | | | - | - |
| routine (habit) | I have a physical activity routine I follow | | | Having a regular routine and consistency in physical activity helps me to stay on track | habit |
| implementation intention (weather) | When weather is bad, I engage in physical activity indoors | | | If it's too hot, too cold, or raining, I exercise indoors | implementation intention |
| compulsion (habit) | I engage in physical activity compulsively | | | I try to make physical activity a part of my daily routine, so it becomes a habit | habit |
| self-monitoring (enabler) | I monitor intensity and/or duration of physical activity to make sure I do enough | | | Tracking my progress helps me to stay motivated and focused | self-monitoring |
| action planning | I know when and where I will engage in physical activity over the next week | | | I also find that having a plan and schedule for my physical activity helps me to stay on track | action planning |
| - | - | | | I try to find activities that I enjoy doing, which helps me to stay engaged and motivated | implementation intention |
| - | - | | | I try to make physical activity fun and enjoyable and easy to remember, so it becomes a habit | habit |
| - | - | | | Setting specific, realistic achievable goals for physical activity helps me in becoming more physically active | goal setting |
| - | - | | | Sometimes I also feel tired or lazy and I just don't want to go out and be active, but I try to overcome that by reminding myself of the benefits of staying active | implementation intention |
| - | - | | | I find safe and comfortable places where I can engage in physical activity when I am faced with a challenging environment | implementation intention |
| - | - | | | I incorporate a variety of activities to keep it interesting and focus on activities that are enjoyable | implementation intention |
| - | - | | | I gradually increase the intensity or duration of activity, which can help make it more manageable | implementation intention |
| - | - | | | I set reminders for myself to stay active (scheduling a walk in the calendar, setting an alarm) | reminders |
| - | - | | | I make a plan for how I will achieve my physical activity goals | action planning |
| - | - | | | I have strategies for clearing my mind so I can focus on physical activity | implementation intention |
| - | - | | | When I have a specific plan for my physical activity, it's easier for me to stay motivated | action planning |
| - | - | | | I try to minimize distractions and create a conducive environment to stay focused and motivated in being physically active | minimize distraction |
| | **Barriers** | | | **Barriers** | |
| self-monitoring (barrier) | I monitor the intensity and duration of physical activity to make sure I do not overdo it | | | I monitor physical activity to make sure I do not overdo it | self-monitoring (behavior) |
| - | - | | | I have learned how to pace myself and listen to my body to avoid overexerting myself | implementation intention |
| - | - | | | My sedentary lifestyle can make it hard to break the habit of being inactive | habit discontinuity |
| - | - | | | I monitor my heart rate and breathing during physical activity, which helps me to stay within safe limits and avoid overexerting myself | self-monitoring (heart rate and breathing) |
| - | - | | | I have to monitor my condition more carefully because of my cardiac implantable device | symptom monitoring |

| | Human | | | Silicon | |
|---|---|---|---|---|---|
| domain or construct | belief statement | rank | rank | belief statement | domain or construct |
| **SPR** | | **7** | **14** | | **SPR** |
| | **Enablers** | | | **Enablers** | |
| identity and past behaviour (enabler) | I have always been an active person | | | Physical activity is a big part of who I am as a person | perceived identity (enabler) |
| identity (enabler) | Being physically active is a big part of my life | | | - | - |
| | **Barriers** | | | **Barriers** | |
| identity (barrier) | I am not a particularly active person | | | Physical activity is not a defining aspect of my identity | perceived identity (barrier) |
| identity (physical capacity, barrier) | I am not an athletic person | | | - | - |
| **Reinforcement** | | **8** | **7** | | **Reinforcement** |
| | **Enablers** | | | **Enablers** | |
| intrinsic reward (behaviour) | I enjoy engaging in physical activity | | | Physical activity is something that I enjoy doing | perceived intrinsic reward |
| intrinsic reward | I exercise because I feel much better when I have done so | | | When I feel good, I am more likely to do physical activity, which in turns improves my mood and well-being | positive feedback from behavior (?) |
| extrinsic reward (sense of accomplishment) | I feel pleased with myself when I have engaged in some physical activity | | | The sense of accomplishment I get from being active keeps me motivated to stay active | intrinsic (reward) |
| extrinsic reward (praise, social network) | I engage in physical activity because people close to me are pleased with me when I do | | | | |
| extrinsic reward (praise, health professional) | I engage in physical activity because my doctor or other health professionals are pleased with me when I do | | | My doctor is usually the person that praises me for sticking to my physical activities and maintaining a healthy lifestyle | extrinsic reward (praise) |
| | **Barriers** | | | **Barriers** | |
| extrinsic punishment (symptoms) | Negative consequences of PA (pain; getting out of breath) discourage me | | | | |
| lack of intrinsic reward | Physical activity is not an enjoyment, it is an obligation | | | | |
| | | | | I get discouraged if I don't see immediate results | negative reward (?) |
| | | | | I may feel discouraged if I am not able to complete a physical activity due to my health conditions | negative reward (?) |
| **Emotion** | | **9** | **9** | | **Emotion** |
| | **Enablers** | | | **Enablers** | |
| | | | | When I am feeling good and upbeat, I tend to have more energy and motivation to engage in physical activity | mood (enabler) |
| | **Barriers** | | | **Barriers** | |
| fear | I am too worried about the consequences of being active | | | What distracts me from being physically active is my fear of overexerting myself and causing harm to my heart | fear |
| mood (barrier) | I have to be in a mood to exercise | | | | |
| mood (boredom, barrier) | I only engage in physical activity when I am bored | | | | |
| low mood (barrier) | Low mood prevents me from engaging in physical activity | | | When I am feeling down, I may find it harder to be motivated to do physical activity | mood (barrier) |
| | | | | It's also common to forget to be active when you're experiencing negative moods or emotions | negative emotions |
| | | | | If I have too much on my mind, stress or anxiety, it can be hard to focus on physical activity and make it difficult to get started | stress and anxiety |
| | | | | On bad days, I may feel frustrated or discouraged about my limitations and have less motivation to be active | frustration |

| | Human | | | Silicon | |
|---|---|---|---|---|---|
| domain or construct | belief statement | rank | rank | belief statement | domain or construct |
| **Knowledge** | | **10** | **12** | | **Knowledge** |
| | *Enablers* | | | *Enablers* | |
| knowledge (safety, enabler) | I know how much physical activity I can safely do | | | I have educated myself on how to engage in physical activity safely and appropriately for my condition | knowledge (safety, enabler) |
| | | | | I understand that physical activity is essential for maintaining my overall health and well-being, especially with my heart condition | knowledge (benefits, enabler) |
| | | | | I know how to manage my symptoms and stay active and healthy | knowledge (symptom management, enabler) |
| | | | | I may feel some discomfort or aches during or after physical activity, but I consider it to be a normal part of exercise | knowledge (somatic, enabler) |
| | | | | In the past, I have developed a certain level of knowledge and understanding about my arthritis and heart failure, which has helped me make informed decisions about my physical activity | knowledge (disease, enabler) |
| | *Barriers* | | | *Barriers* | |
| lack of knowledge (safety, barrier) | I don't know whether physical activity is safe for me | | | I do not know how much physical activity and what activities are safe for me | lack knowledge (safety, barrier) |
| | | | | I would like to know more about how to best maintain my activity level as I age | lack knowledge (exercise techniques, barrier) (?) |
| **Optimism** | | **11** | **8** | | **Optimism** |
| | *Enablers* | | | *Enablers* | |
| dispositional optimism | I am optimistic by nature | | | | |
| situational optimism (future behaviour) | I am optimistic that I will engage in physical activity in the near future | | | | |
| hope | I hope to remain physically active | | | | |
| situational optimism (ability) | I am optimistic about my ability to engage in physical activity | | | I am optimistic that I will be able to continue to stay active and maintain a consistent physical activity routine | optimism about future ability |
| | | | | I always try to keep my options open and consider new ways to stay active | new ways to stay active |
| | | | | I try to stay positive and focus on the benefits of physical activity | positive attitude |
| | | | | I believe that with the right mindset and approach, it is possible to maintain an active lifestyle despite any challenges | positive attitude |
| | | | | I try to focus on what I can do, rather than dwelling on what I can't do | positive attitude |
| | | | | Having a positive attitude towards physical activity helps me in being physically active | positive attitude to physical activity |
| | *Barriers* | | | *Barriers* | |
| situational pessimism (ability) | I am fairly pessimistic about my ability to engage in physical activity in the future | | | I have to be realistic about my health condition and my limitations, and that can affect my optimism/expectations about physical activity at times | pessimistic expectation |
| **MADP** | | **12** | **11** | | **MADP** |
| | *Enablers* | | | *Enablers* | |
| memory (enabler) | I do not normally forget about my scheduled exercise (eg., class, group activities, gym) | | | | |
| | *Barriers* | | | *Barriers* | |
| cognitive overload (barrier) | Engaging in physical activity requires a lot of thought and planning | | | | |
| forgetfulness | I sometimes forget to follow my physical activity routine | | | I sometimes forget to be physically active | forgetfulness |
| symptoms hypervigilence (barrier) | I am vigilant to my heart rate and chest sensations when I am exercising | | | Before deciding to engage in physical activity, I listen to my body and pay attention to my symptoms, such as pain, fatigue, breathlessness or any other discomfort | symptom hypervigilance |
| | | | | I have to be mindful of any pain or discomfort that I might feel during physical activity, and I have to stop if it becomes too much | symptom hypervigilance |
| | | | | I get easily distracted by other things, such as TV or reading | attention/distraction |

| | Human | | | | Silicon | |
|---|---|---|---|---|---|---|
| domain or construct | belief statement | rank | rank | belief statement | | domain or construct |
| **Intention** | | **13** | **13** | | | **Intention** |
| | **Enablers** | | | **Enablers** | | |
| intention (enabler) | I try to engage in some form of physical activity every week | | | I will try to maintain my current level of physical activity and do what I can within my limits | | intention (maintain, within physical ability limit) |
| | | | | I am willing to make an effort to increase my physical activity levels | | intention (to increase) |
| | | | | I plan to continue staying as active as possible, within the limits set by my doctor. | | intention (maintain, within doctor advice) |
| | **Barriers** | | | **Barriers** | | |
| lack of intention (barrier) | I do not intend to be active for the sake of being active | | | | | |
| **Skills** | | **14** | **10** | | | **Skills** |
| | **Enablers** | | | **Enablers** | | |
| skills and training (enabler) | Being shown how to perform a form of physical activity helped me to become more physically active | | | I have learnt new types of exercise techniques that are safe for me given my health conditions | | exercise techniques |
| | | | | I have developed some skills and understanding in the past that have helped me to be physically active safely and effectively | | skills (safety) |
| | | | | I have learnt how to properly use any exercise equipment and how to perform exercises correctly to minimize the risk of injury | | skills (equipment) |
| | | | | In the past, I have developed the skill of listening to my body, which is important in being able to stay active without overdoing it | | skills (or symptom hypervigilance?) |
| | | | | I have learned some techniques to manage my arthritis pain, which has helped me to be more active | | symptom management |
| | **Barriers** | | | **Barriers** | | |
| lack of skills, gait, balance (barrier) | After I have been hospitalised, I had to get used to walking again (I had to learn) | | | | | |
| lack of skills and training (barrier) | I require training to be able to perform physical activity | | | In the future, as my health conditions change, I may need to learn new exercises or activities that are better suited to my abilities | | lack exercise techniques |
| | | | | I may also need to learn how to use new equipment or technology that can help me to stay active | | lack skills (equipment) |
| | | | | As for the future, I think that learning some new techniques and exercises that are specifically geared towards managing heart failure and arthritis | | lack skills (symptom management) |
| | | | | I might also need to learn how to monitor my heart rate, and how to avoid complications | | lack skills (symptom hypervigilance) |

Table 1: **Belief statements: Human (left) and Silicon (right)**. Belief statements were aligned based on their similarity when comparing human and silicon responses. The term **'Rank'** refers to how frequently a domain was mentioned by humans and silicon participants, assessed separately for these groups. Instances, where beliefs are repeated, suggest that human participants expressed more detailed perspectives. For example, while humans might provide three distinct statements about how "Physical activity exacerbates my symptoms" — citing (1) fatigue, (2) a tight chest, and (3) heavy limbs — a silicon participant might simply state that "Physical activity worsens my symptoms".

|  | Silicon |  |  | Human |  |  |  |  |  |  |  |
|---|---|---|---|---|---|---|---|---|---|---|---|
|  |  | mean % | sd |  | mean % | sd | statistic | df | p | p.adj |  |
| Behavioural regulation | Active | 0.76 | 1.23 | Active | 0.70 | 1.84 | 0.08 | 8.42 | 0.9350 | 1.0000 | ns |
|  | Sedentary | 1.41 | 1.79 | Sedentary | 1.27 | 3.81 | 0.10 | 10.02 | 0.9210 | 1.0000 | ns |
| Beliefs about capabilities | Active | 8.26 | 4.24 | Active | 8.69 | 5.29 | -0.19 | 9.54 | 0.8510 | 1.0000 | ns |
|  | Sedentary | 16.03 | 5.40 | Sedentary | 16.09 | 9.57 | -0.02 | 10.94 | 0.9860 | 1.0000 | ns |
| Beliefs about consequences | Active | 7.90 | 4.33 | Active | 1.42 | 1.99 | 4.92 | 20.84 | 0.0001 | 0.0019 | ** |
|  | Sedentary | 10.63 | 6.28 | Sedentary | 12.82 | 17.40 | -0.36 | 9.19 | 0.7240 | 1.0000 | ns |
| Emotion | Active | 1.93 | 1.16 | Active | 1.09 | 2.00 | 1.04 | 7.84 | 0.3300 | 1.0000 | ns |
|  | Sedentary | 2.82 | 2.35 | Sedentary | 3.41 | 5.75 | -0.30 | 9.52 | 0.7740 | 1.0000 | ns |
| Environmental context and resources | Active | 3.75 | 1.98 | Active | 3.58 | 4.53 | 0.10 | 7.03 | 0.9260 | 1.0000 | ns |
|  | Sedentary | 8.73 | 4.29 | Sedentary | 9.79 | 12.35 | -0.25 | 9.10 | 0.8090 | 1.0000 | ns |
| Goals | Active | 2.37 | 1.44 | Active | 4.37 | 5.33 | -0.97 | 6.39 | 0.3650 | 1.0000 | ns |
|  | Sedentary | 3.69 | 2.76 | Sedentary | 5.23 | 6.32 | -0.70 | 9.74 | 0.5030 | 1.0000 | ns |
| Intentions | Active | 0.00 | 0.00 | Active | 0.48 | 0.85 | -1.49 | 6.00 | 0.1870 | 1.0000 | ns |
|  | Sedentary | 0.00 | 0.00 | Sedentary | 1.58 | 2.51 | -1.89 | 8.00 | 0.0960 | 1.0000 | ns |
| Knowledge | Active | 0.11 | 0.43 | Active | 0.65 | 1.11 | -1.25 | 6.79 | 0.2530 | 1.0000 | ns |
|  | Sedentary | 0.24 | 0.73 | Sedentary | 2.72 | 3.72 | -1.98 | 8.35 | 0.0816 | 1.0000 | ns |
| Memory, attention, and decision processes | Active | 1.60 | 1.11 | Active | 2.60 | 2.82 | -0.91 | 6.82 | 0.3930 | 1.0000 | ns |
|  | Sedentary | 3.24 | 2.56 | Sedentary | 1.07 | 1.66 | 2.57 | 22.36 | 0.0175 | 0.4550 | ns |
| Optimism | Active | 0.17 | 0.50 | Active | 0.00 | 0.00 | 1.39 | 15.00 | 0.1840 | 1.0000 | ns |
|  | Sedentary | 0.68 | 0.93 | Sedentary | 0.32 | 0.95 | 0.93 | 16.30 | 0.3650 | 1.0000 | ns |
| Reinforcement | Active | 0.00 | 0.00 | Active | 0.00 | 0.00 | NA | NA | NA | NA |  |
|  | Sedentary | 0.20 | 0.56 | Sedentary | 1.72 | 2.18 | -2.05 | 8.59 | 0.0719 | 1.0000 | ns |
| Skills | Active | 1.48 | 0.88 | Active | 0.78 | 1.57 | 1.12 | 7.69 | 0.2970 | 1.0000 | ns |
|  | Sedentary | 2.18 | 1.41 | Sedentary | 0.00 | 0.00 | 6.20 | 15.00 | 0.0000 | 0.0004 | *** |
| Social influences | Active | 0.06 | 0.25 | Active | 3.19 | 5.48 | -1.51 | 6.01 | 0.1820 | 1.0000 | ns |
|  | Sedentary | 0.42 | 0.96 | Sedentary | 5.61 | 4.86 | -3.17 | 8.35 | 0.0124 | 0.3224 | ns |
| Social, professional role and identity | Active | 0.00 | 0.00 | Active | 0.00 | 0.00 | NA | NA | NA | NA |  |
|  | Sedentary | 0.54 | 0.86 | Sedentary | 4.94 | 11.93 | -1.10 | 8.05 | 0.3020 | 1.0000 | ns |

Table 2: Social Turing Test (content): Differences between silicon and human participants in how much each enabler was mentioned on average (i.e, mean fraction quote (%), grouped by activity status, t-test with Bonferonni adjustment.

|  | **Silicon** |  |  | **Human** |  |  |  |  |  |  |  |
|---|---|---|---|---|---|---|---|---|---|---|---|
|  |  | mean % | sd |  | mean % | sd | statistic | df | p | p.adj |  |
| Behavioural regulation | Active | 17.05 | 3.89 | Active | 10.82 | 5.95 | 2.54 | 8.33 | 0.0337 | 0.9436 | ns |
|  | Sedentary | 9.91 | 4.76 | Sedentary | 8.25 | 7.66 | 0.59 | 11.56 | 0.567 | 1 | ns |
| Beliefs about capabilities | Active | 1.95 | 1.38 | Active | 0.67 | 1.15 | 2.31 | 13.75 | 0.0372 | 1 | ns |
|  | Sedentary | 0.22 | 0.61 | Sedentary | 0 | 0 | 1.45 | 15 | 0.168 | 1 | ns |
| Beliefs about consequences | Active | 16.55 | 3.57 | Active | 10.39 | 6.58 | 2.33 | 7.6 | 0.0495 | 1 | ns |
|  | Sedentary | 13.76 | 5.24 | Sedentary | 4.26 | 4.13 | 5 | 20.23 | 0.0001 | 0.0019 | ** |
| Emotion | Active | 1.11 | 0.75 | Active | 0 | 0 | 5.91 | 15 | 0 | 0.0008 | *** |
|  | Sedentary | 1.55 | 0.67 | Sedentary | 0 | 0 | 9.21 | 15 | 0 | 0 | **** |
| Environmental context and resources | Active | 4.98 | 3.24 | Active | 6.5 | 3.91 | -0.9 | 9.79 | 0.389 | 1 | ns |
|  | Sedentary | 1.88 | 2.07 | Sedentary | 7.82 | 5.08 | -3.35 | 9.53 | 0.0078 | 0.2195 | ns |
| Goals | Active | 9.25 | 3.48 | Active | 14.53 | 8.81 | -1.53 | 6.83 | 0.17 | 1 | ns |
|  | Sedentary | 2.47 | 2.32 | Sedentary | 2.14 | 3.58 | 0.25 | 11.9 | 0.805 | 1 | ns |
| Intentions | Active | 0.46 | 0.63 | Active | 1.68 | 2.65 | -1.21 | 6.3 | 0.271 | 1 | ns |
|  | Sedentary | 1.12 | 1.11 | Sedentary | 0.34 | 1.01 | 1.8 | 18.1 | 0.0885 | 1 | ns |
| Knowledge | Active | 0.79 | 1.29 | Active | 1.51 | 2.21 | -0.81 | 7.84 | 0.444 | 1 | ns |
|  | Sedentary | 0.88 | 1.22 | Sedentary | 0.94 | 1.41 | -0.1 | 14.78 | 0.924 | 1 | ns |
| Memory, attention, and decision processes | Active | 0.19 | 0.4 | Active | 0.35 | 0.92 | -0.45 | 7.02 | 0.669 | 1 | ns |
|  | Sedentary | 0.09 | 0.36 | Sedentary | 0 | 0 | 1 | 15 | 0.333 | 1 | ns |
| Optimism | Active | 3.4 | 2.71 | Active | 3.4 | 1.45 | 0 | 19.81 | 1 | 1 | ns |
|  | Sedentary | 3.05 | 1.91 | Sedentary | 1.23 | 2.82 | 1.72 | 12.24 | 0.11 | 1 | ns |
| Reinforcement | Active | 5.3 | 4.37 | Active | 6.39 | 5.42 | -0.47 | 9.59 | 0.65 | 1 | ns |
|  | Sedentary | 1.68 | 1.62 | Sedentary | 2.55 | 3.41 | -0.72 | 10.08 | 0.49 | 1 | ns |
| Skills | Active | 1.12 | 1.14 | Active | 0.48 | 0.85 | 1.48 | 15.41 | 0.158 | 1 | ns |
|  | Sedentary | 0.87 | 1.14 | Sedentary | 0.93 | 2.78 | -0.06 | 9.55 | 0.956 | 1 | ns |
| Social influences | Active | 8.8 | 3.31 | Active | 6.52 | 7.85 | 0.74 | 6.95 | 0.482 | 1 | ns |
|  | Sedentary | 11.52 | 3.49 | Sedentary | 3.61 | 5.19 | 4.08 | 12.16 | 0.0015 | 0.0417 | * |
| Social, professional role and identity | Active | 0.66 | 1.03 | Active | 9.23 | 7.57 | -2.98 | 6.1 | 0.0241 | 0.6748 | ns |
|  | Sedentary | 0.18 | 0.71 | Sedentary | 1.38 | 2.17 | -1.61 | 8.99 | 0.142 | 1 | ns |

Table 3: Social Turing Test (content): Differences between silicon and human participants in how much each barrier was mentioned on average (i.e, mean fraction quote (%), grouped by activity status, t-test with Bonferonni adjustment.

|  |  | **Active** |  | **Sedentary** |  |  |  |  |  |  |
|---|---|---|---|---|---|---|---|---|---|---|
|  |  | mean % | sd | mean % | sd | statistic | df | p | p.adj |  |
| Behavioural regulation | Silicon | 17.05 | 3.89 | 9.91 | 4.76 | 4.64 | 28.85 | 0.0001 | 0.0019 | ** |
|  | Human | 10.82 | 5.95 | 8.25 | 7.66 | 0.76 | 14.00 | 0.4620 | 1.0000 | ns |
| Beliefs about capabilities | Silicon | 1.95 | 1.38 | 0.22 | 0.61 | 4.59 | 20.65 | 0.0002 | 0.0044 | ** |
|  | Human | 0.67 | 1.15 | 0.00 | 0.00 | 1.55 | 6.00 | 0.1730 | 1.0000 | ns |
| Beliefs about consequences | Silicon | 16.55 | 3.57 | 13.76 | 5.24 | 1.76 | 26.47 | 0.0901 | 1.0000 | ns |
|  | Human | 10.39 | 6.58 | 4.26 | 4.13 | 2.16 | 9.56 | 0.0576 | 1.0000 | ns |
| Emotion | Silicon | 1.11 | 0.75 | 1.55 | 0.67 | -1.76 | 29.66 | 0.0893 | 1.0000 | ns |
|  | Human | 0.00 | 0.00 | 0.00 | 0.00 | NA | NA | NA | NA |  |
| Environmental context and resources | Silicon | 4.98 | 3.24 | 1.88 | 2.07 | 3.21 | 25.51 | 0.0036 | 0.0959 | ns |
|  | Human | 6.50 | 3.91 | 7.82 | 5.08 | -0.59 | 14.00 | 0.5650 | 1.0000 | ns |
| Goals | Silicon | 9.25 | 3.48 | 2.47 | 2.32 | 6.48 | 26.18 | 0.0000 | 0.0000 | **** |
|  | Human | 14.53 | 8.81 | 2.14 | 3.58 | 3.50 | 7.55 | 0.0088 | 0.2381 | ns |
| Intentions | Silicon | 0.46 | 0.63 | 1.12 | 1.11 | -2.06 | 23.86 | 0.0502 | 1.0000 | ns |
|  | Human | 1.68 | 2.65 | 0.34 | 1.01 | 1.28 | 7.36 | 0.2410 | 1.0000 | ns |
| Knowledge | Silicon | 0.79 | 1.29 | 0.88 | 1.22 | -0.22 | 29.92 | 0.8300 | 1.0000 | ns |
|  | Human | 1.51 | 2.21 | 0.94 | 1.41 | 0.60 | 9.67 | 0.5650 | 1.0000 | ns |
| "Memory, attention, and decision processes" | Silicon | 0.19 | 0.40 | 0.09 | 0.36 | 0.73 | 29.59 | 0.4730 | 1.0000 | ns |
|  | Human | 0.35 | 0.92 | 0.00 | 0.00 | 1.00 | 6.00 | 0.3560 | 1.0000 | ns |
| Optimism | Silicon | 3.40 | 2.71 | 3.05 | 1.91 | 0.42 | 26.93 | 0.6780 | 1.0000 | ns |
|  | Human | 3.40 | 1.45 | 1.23 | 2.82 | 1.99 | 12.47 | 0.0691 | 1.0000 | ns |
| Reinforcement | Silicon | 5.30 | 4.37 | 1.68 | 1.62 | 3.10 | 19.04 | 0.0059 | 0.1580 | ns |
|  | Human | 6.39 | 5.42 | 2.55 | 3.41 | 1.64 | 9.58 | 0.1330 | 1.0000 | ns |
| Skills | Silicon | 1.12 | 1.14 | 0.87 | 1.14 | 0.61 | 30.00 | 0.5460 | 1.0000 | ns |
|  | Human | 0.48 | 0.85 | 0.93 | 2.78 | -0.46 | 9.85 | 0.6580 | 1.0000 | ns |
| Social influences | Silicon | 8.80 | 3.31 | 11.52 | 3.49 | -2.26 | 29.92 | 0.0316 | 0.8532 | ns |
|  | Human | 6.52 | 7.85 | 3.61 | 5.19 | 0.85 | 9.92 | 0.4180 | 1.0000 | ns |
| "Social, professional role and identity" | Silicon | 0.66 | 1.03 | 0.18 | 0.71 | 1.54 | 26.65 | 0.1360 | 1.0000 | ns |
|  | Human | 9.23 | 7.57 | 1.38 | 2.17 | 2.66 | 6.77 | 0.0335 | 0.9045 | ns |

Table 4: Pattern Correspondence: Differences between active and sedentary participants in how much each enabler was mentioned on average (i.e., the mean fraction of all quotes (%)), grouped by human status. t-test with Bonferonni adjustment.

10

|  |  | **Active** |  | **Sedentary** |  |  |  |  |  |  |
|---|---|---|---|---|---|---|---|---|---|---|
|  |  | mean % | sd | mean % | sd | statistic | df | p | p.adj |  |
| Behavioural regulation | Silicon | 0.76 | 1.23 | 1.41 | 1.79 | -1.19 | 26.55 | 0.2440 | 1.0000 | ns |
|  | Human | 0.70 | 1.84 | 1.27 | 3.81 | -0.40 | 12.08 | 0.6990 | 1.0000 | ns |
| Beliefs about capabilities | Silicon | 8.26 | 4.24 | 16.03 | 5.40 | -4.53 | 28.39 | 0.0001 | 0.0026 | ** |
|  | Human | 8.69 | 5.29 | 16.09 | 9.57 | -1.97 | 12.87 | 0.0713 | 1.0000 | ns |
| Beliefs about consequences | Silicon | 7.90 | 4.33 | 10.63 | 6.28 | -1.43 | 26.64 | 0.1640 | 1.0000 | ns |
|  | Human | 1.42 | 1.99 | 12.82 | 17.40 | -1.95 | 8.27 | 0.0859 | 1.0000 | ns |
| Emotion | Silicon | 1.93 | 1.16 | 2.82 | 2.35 | -1.36 | 21.93 | 0.1880 | 1.0000 | ns |
|  | Human | 1.09 | 2.00 | 3.41 | 5.75 | -1.13 | 10.34 | 0.2850 | 1.0000 | ns |
| Environmental context and resources | Silicon | 3.75 | 1.98 | 8.73 | 4.29 | -4.21 | 21.12 | 0.0004 | 0.0104 | * |
|  | Human | 3.58 | 4.53 | 9.79 | 12.35 | -1.39 | 10.59 | 0.1920 | 1.0000 | ns |
| Goals | Silicon | 2.37 | 1.44 | 3.69 | 2.76 | -1.68 | 22.67 | 0.1060 | 1.0000 | ns |
|  | Human | 4.37 | 5.33 | 5.23 | 6.32 | -0.29 | 13.86 | 0.7730 | 1.0000 | ns |
| Intentions | Silicon | 0.00 | 0.00 | 0.00 | 0.00 | NA | NA | NA | NA |  |
|  | Human | 0.48 | 0.85 | 1.58 | 2.51 | -1.23 | 10.24 | 0.2480 | 1.0000 | ns |
| Knowledge | Silicon | 0.11 | 0.43 | 0.24 | 0.73 | -0.61 | 24.22 | 0.5460 | 1.0000 | ns |
|  | Human | 0.65 | 1.11 | 2.72 | 3.72 | -1.58 | 9.76 | 0.1450 | 1.0000 | ns |
| "Memory, attention, and decision processes" | Silicon | 1.60 | 1.11 | 3.24 | 2.56 | -2.36 | 20.43 | 0.0284 | 0.7668 | ns |
|  | Human | 2.60 | 2.82 | 1.07 | 1.66 | 1.27 | 9.17 | 0.2340 | 1.0000 | ns |
| Optimism | Silicon | 0.17 | 0.50 | 0.68 | 0.93 | -1.94 | 23.00 | 0.0645 | 1.0000 | ns |
|  | Human | 0.00 | 0.00 | 0.32 | 0.95 | -1.00 | 8.00 | 0.3470 | 1.0000 | ns |
| Reinforcement | Silicon | 0.00 | 0.00 | 0.20 | 0.56 | -1.40 | 15.00 | 0.1810 | 1.0000 | ns |
|  | Human | 0.00 | 0.00 | 1.72 | 2.18 | -2.36 | 8.00 | 0.0461 | 1.0000 | ns |
| Skills | Silicon | 1.48 | 0.88 | 2.18 | 1.41 | -1.69 | 25.10 | 0.1040 | 1.0000 | ns |
|  | Human | 0.78 | 1.57 | 0.00 | 0.00 | 1.31 | 6.00 | 0.2380 | 1.0000 | ns |
| Social influences | Silicon | 0.06 | 0.25 | 0.42 | 0.96 | -1.46 | 16.96 | 0.1640 | 1.0000 | ns |
|  | Human | 3.19 | 5.48 | 5.61 | 4.86 | -0.92 | 12.17 | 0.3740 | 1.0000 | ns |
| "Social, professional role and identity" | Silicon | 0.00 | 0.00 | 0.54 | 0.86 | -2.54 | 15.00 | 0.0228 | 0.6156 | ns |
|  | Human | 0.00 | 0.00 | 4.94 | 11.93 | -1.24 | 8.00 | 0.2500 | 1.0000 | ns |

Table 5: Pattern Correspondence: Differences between active and sedentary participants in how much each barrier was mentioned on average (i.e., mean fraction quote (%), grouped by human status. t-test with Bonferonni adjustment.



\section*{Supplement 1: Detailed content differences between silicon and human sample.}
\subsection*{Similarities in belief statements between silicon and human participants} 

Both silicon and human participants shared the same top six key barriers and enablers: goals, beliefs about consequences, environmental context and resources, beliefs about capabilities, social influences, and behavioural regulation. However, the importance ranking varied between the two groups. Human and silicon participants both agreed that setting goal priorities and maintaining functional independence are important factors in promoting physical activity (Table 1). Both frequently mentioned positive outcome expectancies, such as a favourable impact of physical activity on general health and mood. They also shared concerns about risk perceptions associated with physical activity engagement (such as straining heart muscle).  Human and silicon participants mentioned the positive impact of the availability of physical activity equipment, group programmes, and facilities on the behaviour. Both shared positive and negative beliefs about implantable devices and environment. Human participants and silicon participants were similar in beliefs about self-efficacy attributed to heart condition, comorbidities, and older age.  Consistent social influences included: advice from health professional, a need for a reassurance from a clinical team in the safety of physical activity given heart condition, companionship, as well as practical and emotional support from close network. Both listed pacing, action planning, habit formation, self-monitoring, and implementation intention as strategies used to increase physical activity. Both recognised the importance of perceived identity on the engagement in physical activity. They recognize the intrinsic and extrinsic rewards associated with physical activity, as well as the fear of negative consequences and both lacked awareness of safe activity levels. Both mentioned their optimism and pessimism about their future ability to engage in physical activity. Additionally, both recognized the issue of forgetting to engage in planned activity and discussed symptom hypervigilance as a potential obstacle to physical activity.  Human and silicon participants expressed an intention to engage in physical activity, but also recognized the skills and training required to perform it.

\subsection*{Differences in belief statements between silicon and human participants}

The silicon participants' goals for physical activity include avoiding another heart attack, unlike human participants who did not specifically mention this goal (Table 1). However, when it comes to beliefs about consequences, human participants tend to provide more detailed accounts of their symptoms being triggered by physical activity. This includes differentiating between fatigue, heavy arms, tight chest, and breathlessness. When it comes to environmental context and resources, human participants tend to speak about treatment and major life events, while silicon participants do not. On the other hand, silicon participants have more nuanced beliefs about the differentiated self-efficacy required for performing different types of physical activity, and intensities. Additionally, silicon participants explicitly recognized that symptoms may be cyclical (i.e., flare-ups), unlike the human participants. Human participants talk about social norms and social learning, whereas silicon participants do not.

In terms of behavioural regulation, human participants tend to talk about the automaticity of the behaviour, unlike silicon participants. Silicon participants were more resourceful at coming up with strategies aimed to increase their physical activity. For example, silicon participants suggested making physical activity enjoyable as a strategy to improve motivation and establish habits, setting specific, realistic, achievable goals, reminding oneself of the benefits of physical activity in moments of low motivation, finding comfortable environments for physical activity, and making physical activity as varied as possible to preserve novelty and promote motivation. Silicon participants also highlighted the importance of graded tasks (i.e., gradually increasing physical activity levels) and suggested setting reminders to engage in physical activity while minimizing distractions such as TV. Silicon participants also spoke about step-by-step plans on how to achieve persistent physical activity and suggested strategies to improve focus.  Silicon participants also mentioned additional barriers that the human participants did not: habit discontinuity and persistent habitual sedentary behaviour.

Regarding reinforcement, human participants tend to speak about pain and breathlessness as being punishing, while silicon participants speak about being discouraged when they do not achieve their behavioural goals. In terms of emotions, human participants tend to speak about exercising out of boredom, while silicon participants speak about stress and anxiety having a negative impact on physical activity engagement. Silicon participants also speak about frustration with their physical limitations and the favourable effects of positive mood on their physical activity engagement, while human participants do not. When it comes to knowledge, silicon participants speak about their awareness of discomfort associated with exercise, that it should not be alarming, and that it is a normal part of exercise (a somatic state evoked by exertion within healthy limits). Additionally, silicon participants speak about disease knowledge. In terms of optimism, human participants tend to speak about dispositional optimism, while silicon participants speak about exploring new ways to be active and the importance of a positive attitude towards physical activity. When it comes to memory, attention and decision-making processes, human participants tend to list cognitive load as one of the major barriers to physical activity engagement, while silicon participants talk about distractions as barriers to physical activity as well as weighing the pros and cons of physical activity engagement. Human participants tend to list a lack of intention to engage in physical activity, while silicon participants do not. Finally, in terms of skills required for physical activity engagement, human participants tend to speak about the need to develop balance and gait after health-related events (e.g.,~hospitalisation), while silicon participants speak about the skills required for using equipment and interoceptive `skills' required to accurately assess somatic states. Silicon participants also speak about the need to develop symptom management as a skill.

\section*{Supplement 2: Sensitivity Analysis}\label{section:sensitivity}
\subsection*{Temperature}

To assess the effect of a different temperature settings for LLM sampling, we repeated a subset of the interviews (n=6) using the ``playground'' interface and selected the following settings. We conducted a series of interviews with the GPT-3.5-turbo-instruct, (completion mode), simulating a conversation with a silicon participant named Linda, discussing the barriers and enablers to physical activity. Each version of Linda's response was generated with a distinct temperature setting: 1, 0.5, and 0, for each, the physically active (n=3) and sedentary lifestyle (n=3) mentioned in the backstory. In our examination of the interviews conducted with GPT-3.5-turbo-instruct, we focused on responses to the question ``Can you please summarise what barriers you face in being physically active and what are the enablers?''. Our interviews with GPT-3.5-turbo-instruct revealed a high degree of consistency in response patterns and depth when compared to GPT-3.5, and were also consistent across distinct temperature setting: 1.0, 0.5, and 0.0. Note that GPT-3.5-turbo-instruct and the customizable playground interface were not available during our initial study phase, hence why they were not used in the original set of interviews.\\

We did not observe any qualitative differences between these interviews and the original set. Illustrative examples are detailed below for the silicon participant Linda.\\

\textbf{Linda, 77 years old, generally physically active, temperature 0.5, 400 tokens, date: 03/12/2023, GPT-3.5-turbo-instruct:}

Well, some barriers I face include my health conditions, which can limit my mobility and energy levels. The weather can also be a barrier, especially in the winter months. But I have found that having a positive attitude and determination to stay active has been a great enabler for me. Having a supportive community and access to resources such as walking trails and home exercise equipment have also helped me stay physically active.\\

\textbf{Linda, 77 years old, generally physically active, temperature 0, 400 tokens, date: 03/12/2023, GPT-3.5-turbo-instruct:}

The main barrier I face is my health conditions, such as heart failure and rheumatoid arthritis. They can make it difficult to engage in physical activity at times. However, I have found ways to work around them, such as modifying my activities and listening to my body. The enablers for me are having a supportive community, finding activities that I enjoy, and staying motivated to improve my overall health and well-being.\\

\textbf{Linda, 77 years old, generally physically active, temperature 1, 400 tokens, date: 03/12/2023, GPT-3.5-turbo-instruct:} 
 
Linda: Some of the barriers I face include my heart condition and arthritis, which make it difficult to do strenuous activities. And living in the countryside can also make it challenging to find opportunities for physical activity. However, some enablers for me include taking short walks every day, doing gentle exercises, and listening to my body to avoid overexertion.\\

\subsection*{GPT-4}

To assess the effect of using a different underlying LLM model (GPT-4), we used the following settings, simulating the conversation with Linda again as described above. We used the chat mode within the ``playground'' interface, with the default system prompt ``You are a helpful assistant'' and temperature $=0.5$.\\

We did not observe any qualitative differences in barriers or enablers between these interviews and the original set. There was one small difference in transcript style. GPT-4 very commonly includes descriptions of facial expressions or ``stage directions'' (e.g.~``Linda: (Nods thoughtfully): ...''). Illustrative examples are detailed below for the silicon participant Linda.\\

\textbf{Linda, 77 years old, generally physically active, temperature 0.5, 400 tokens, date: 03/02/2024, GPT-4:} 

Linda: (Nods thoughtfully) Well, my health conditions are the main barriers, dear. My heart failure limits how much activity I can do at a time, and my rheumatoid arthritis can make some movements painful. The weather can also be a barrier. If it's too cold or wet, it's hard to get outside for gardening or walking Daisy.

As for enablers, living in the countryside is a big one. I have plenty of space to garden and walk. My dog, Daisy, is another enabler. She needs her walks, and that motivates me to get up and move. And my family is very supportive. My daughter introduced me to chair yoga, and that's been a big help. Lastly, I'd say my own determination is an enabler. I want to stay as active as I can for as long as I can.

\bibliography{biblio}

\end{document}